\documentclass[final,5p,times,twocolumn,authoryear,nopreprintline]{elsarticle}

\usepackage{geometry} % added 27-02-2014 Markus Englich
\usepackage{amsmath}
\usepackage{comment}
\usepackage{algorithm}
\usepackage[noend]{algpseudocode}
\usepackage{varwidth}% http://ctan.org/pkg/varwidth

\usepackage{amssymb}
\usepackage{subcaption}

\usepackage{xargs}                      % Use more than one optional parameter in a new commands
\usepackage[pdftex,dvipsnames]{xcolor}  % Coloured text etc.
\usepackage[colorinlistoftodos,prependcaption,textsize=tiny]{todonotes}
\newcommandx{\unsure}[2][1=]{\todo[linecolor=red,backgroundcolor=red!25,bordercolor=red,#1]{#2}}
\newcommandx{\change}[2][1=]{\todo[linecolor=blue,backgroundcolor=blue!25,bordercolor=blue,#1]{#2}}
\newcommandx{\info}[2][1=]{\todo[linecolor=OliveGreen,backgroundcolor=OliveGreen!25,bordercolor=OliveGreen,#1]{#2}}
\newcommandx{\improvement}[2][1=]{\todo[linecolor=Plum,backgroundcolor=Plum!25,bordercolor=Plum,#1]{#2}}
\newcommandx{\thiswillnotshow}[2][1=]{\todo[disable,#1]{#2}}

\usepackage{booktabs}
\newcommand{\ra}[1]{\renewcommand{\arraystretch}{#1}}

\usepackage{stfloats}

\usepackage{hyperref}

\journal{ISPRS Journal of Photogrammetry and Remote Sensing}

\begin{document}

\begin{frontmatter}



\title{Instance segmentation of fallen trees in aerial color infrared imagery
using active multi-contour evolution with fully convolutional network-based intensity
priors}


\author[polyu]{Przemyslaw Polewski}
\ead{przemyslaw.polewski@gmail.com}
\author[polyu]{Jacquelyn Shelton}
\ead{jacquelyn.ann.shelton@gmail.com}
\author[polyu]{Wei Yao\corref{cor1}}
\ead{wei.hn.yao@polyu.edu.hk}
\author[npbw]{Marco Heurich}
\ead{marco.heurich@npv-bw.bayern.de}

\address[polyu]{Dept. of Land Surveying and Geo-Informatics,
The Hong Kong Polytechnic University, Hung Hom, Kowloon, Hong Kong SAR, China}
\address[npbw]{Dept. for Visitor Management and National Park Monitoring, Bavarian Forest
	National Park, 94481 Grafenau, Germany}
	
\cortext[cor1]{Corresponding author.}

\vskip 1cm

\begin{abstract}
Over the last several years, semantic image segmentation based on 
deep neural networks has been greatly advanced. On the other hand,
single-instance segmentation still remains a challenging problem. In this paper,
we introduce a framework for segmenting instances of a common object class by
multiple active contour evolution over semantic segmentation maps of images
obtained through fully convolutional networks. The contour evolution is cast as
an energy minimization problem, where the aggregate energy functional
incorporates a data fit term, an explicit shape model, and accounts for object
overlap. Efficient solution neighborhood operators are proposed,
enabling optimization through metaheuristics such as simulated annealing. We
instantiate the proposed framework in the context of segmenting individual
fallen stems from high-resolution aerial multispectral imagery, providing
problem-specific energy potentials. We validated our approach on 3 real-world
scenes of varying complexity, using 730 manually labeled polygon outlines as ground
truth. The test plots were situated in regions of the Bavarian Forest National
Park, Germany, which sustained a heavy bark beetle infestation. Evaluations were
performed on both the polygon and line segment level, showing that the
multi-contour segmentation can achieve up to 0.93 precision and 0.82 recall. An
improvement of up to 7 percentage points (pp) in recall and 6 in precision
compared to an iterative sample consensus line segment detection baseline was
achieved. Despite the simplicity of the applied shape parametrization, an
explicit shape model incorporated into the energy function improved
the results by up to 4 pp of recall. Finally, we show the importance of
using a high-quality semantic segmentation method (e.g.~U-net) as the basis for
individual stem detection, as the quality of the results degraded dramatically
in our baseline experiment utilizing a simpler method. Our method is a step
towards increased accessibility of automatic fallen tree mapping in forests, due
to higher cost efficiency of aerial imagery acquisition
compared to laser scanning. The precise fallen tree maps could be further used
as a basis for plant and animal habitat modeling, studies on carbon
sequestration as well as soil quality in forest ecosystems.

\end{abstract}


\end{frontmatter}



\section{Introduction}\label{sec:introduction}

Forest ecosystems are the most species-rich ecosystems on earth and play
an essential role in providing ecosystem services such as wood production,
drinking water supply, carbon sequestration, and biodiversity
preservation~\citep{watsonExceptionalValueIntact2018}. However, forests are
under immense pressure especially because of the unsustainable use of their
resources, conversion into other land use types, and global change. Therefore,
there is a strong need for better management and conservation practises
allowing a sustainable use that can secure all the services. A critical
precondition for sustainable forest management are monitoring schemes that
provide the necessary information for preparing management plans. Besides
growing stock, yield and tree species distribution, also deadwood is an
essential indicator of forest health as e.g.~in temperate forests, up to
one-third of all species depend on it during their life
cycle~\citep{Mller2010ARO}. Moreover, it is not just the amount of deadwood
that matters for the conservation of biodiversity~\citep{Seibold2018}; also its
quality is decisive for the conservation of biodiversity. Therefore, it is also
important to determine the tree species, the decay stage and if the dead wood
is standing or lying. Estimating the amount of both types of deadwood is not
just decisive for the maintaining biodiversity, but also for management of
adverse effects of forest disturbances such as wind throws and insect
outbreaks. The latter aspect is becoming increasingly important as the
frequency and severity of such disturbance events are continuously on the rise
due to global change~\citep{Seidl2017}. Such events can affect large tracks of
forested land in a short time span (e.g. windthrow). That makes it very
difficult to accurately assess the amount of timber affected by conventional
field-based methods. Therefore, remote sensing techniques are a natural and
cost-efficient alternative to field works. This demand for accurate information
on the distribution of coarse woody debris (CWD) in forests, driven by
the aforementioned factors, has sparked research interest within the remote
sensing community. In recent years, a number of contributions have focused on
detection and classification of dead wood from laser scanning
data~\citep{rs10091356}. Delineating individual fallen stems in both
aerial~(e.g.~\citet{Polewski2015b}) and
terrestrial~(e.g.~\citet{POLEWSKI2017118}) point clouds was shown to be feasible.

While the 3D information inherent in laser scanning point clouds provides a
solid basis for fallen tree detection, obtaining high density laser scanning data may be
prohibitively expensive.
Multispectral aerial imagery offers a more accessible alternative. The
near-infrared channel is particularly useful for this purpose, since dead and
diseased vegetation produces a distinct reflectance signature in this spectral
band~\citep{Jensen06}. In case of fallen tree stems, resolution at the level of
decimeters or better is crucial to the success of detection, because the width
of the target object can be as low as 30-40 cm, and as such they could appear as
only a single row of pixels (or be altogether missing) within a lower-resolution
image. A number of studies considered the determination of tree health
(e.g.~\citet{rs11060643}) or direct detection of coarse woody debris (e.g.~\citet{Freeman2016}) from high-resolution optical imagery. Currently, most approaches dealing with fallen trees focus on either analyzing groupings of pixels without a one-to-one correspondence to stems (e.g.~\citet{f8010021,f10060471}), or determining lines which represent the positions and lengths of individual trees, disregarding their
thickness~\citep{Panagiotidis2019,rs9040306}.
\begin{figure*}[ht]
  \centering
  \begin{subfigure}[b]{0.32\textwidth}
  	\includegraphics[width=\textwidth]{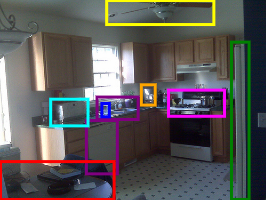}
  	\caption{}
  	\label{fig:closeRange1}
  \end{subfigure}
  \hspace{\fill}
  \begin{subfigure}[b]{0.32\textwidth}
  	\includegraphics[width=\textwidth]{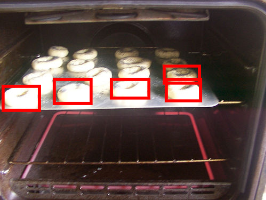}
  	\caption{}
  	\label{fig:closeRange2}
  \end{subfigure}
  \hspace{\fill}
  \begin{subfigure}[b]{0.32\textwidth}
  	\includegraphics[width=\textwidth]{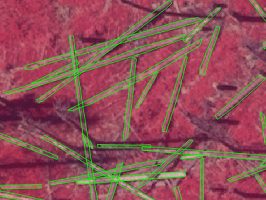}
  	\caption{}
  	\label{fig:sampleFallen}
  \end{subfigure}
  \caption{(a)-(b) sample images from the LSVRC dataset
  \citep{ILSVRC15}. The data is geared towards high-resolution close-range
  photography from handheld devices. (c) sample nadir-view color infrared image
  showing multiple intersecting fallen stems.}
  \label{fig:scenarioStemsSample}
\end{figure*}

This paper considers the task of delineating single fallen stems in the broader
context of instance segmentation in imagery. Our goal is to extract polygons
representing individual stems from difficult scenarios which contain dozens of
partially overlapping and intersecting objects (see
Fig.~\ref{fig:scenarioStemsSample}). While dense semantic segmentation of images
has been arguably all but solved using decoder-encoder architectures like fully
convolutional networks (e.g.~\citet{RFB15a}), extraction of individual object
\emph{instances} still remains a challenge and an active area of research
within the neural network and computer vision
community~\citep{DBLP:journals/corr/ArnabT17,8237584}.
State-of-the-art approaches based on convolutional neural networks(CNN) suffer
from coarseness of feature maps and limited information contained in the
candidate object regions of interest, which leads to degraded performance for
small and multi-scale object
localization~\citep{DBLP:journals/corr/abs-1807-05511}. Also, some of the recent
advancements have been geared towards benchmarks and competitions published by
the computer vision community, such as the Large Scale Visual Recognition
Challenge (LSVRC)~\citep{ILSVRC15}. These datasets usually contain large
quantities of close-range images captured from handheld devices, depicting
clearly-visible 'common' objects such as household items, people, animals, etc.
The emphasis is put on the network's ability to recognize a variety of object
classes (the LSVRC data contains 200 categories). In contrast, remote sensing
images, especially acquired in a natural resource monitoring setting, usually
contain many possibly overlapping instances of the same object category, like
fallen trees in a bark beetle attack zone (Fig.~\ref{fig:scenarioStemsSample})
or a cluster of tree crowns. Although the optical sensor hardware is improving,
the average resolution of aerial remote sensing imagery is still
significantly smaller than in case of close-range photography, resulting in
possibly blurred object boundaries. This could pose a problem to some of the
state-of-the-art CNN architectures which follow a region proposal-classification
design (e.g.~\citet{7485869}). Overlapping region proposals, containing
candidate object bounding boxes, are usually pruned using a discrete process like
non-maxima suppression, which means that if the candidate generator produces a
high 'objectness' score on an image region not centered on a real object (due
to blurred boundaries and heavy candidate overlap), the true detections could be
thrown away and never even make it to the classification stage. Moreover,
specifically for the case of fallen stems, detection based on axis aligned
bounding boxes has a key weakness. Typically, imagery obtained in remote sensing
flight campaigns features a ground sampling distance of no less than 10-15 cm.
Therefore, fallen tree trunks would appear only several pixels wide and
possibly hundreds of pixels long. Assuming that the stems may be arbitrarily
oriented within the image, the tree's axis-aligned bounding box would be
overwhelmingly populated with irrelevant pixels (except close to the main
diagonal).

To alleviate some of these problems, we introduce a general framework for
segmenting sets of overlapping objects of a single category into individual
instances. Instead of attempting to detect objects in axis-aligned bounding
boxes, we maintain shape parametrizations and associated rigid transform
parameters separately per instance, effectively evolving multiple active
contours~\citep{CremersIJCV} simultaneously. Our framework explicitly models
object overlap as well as prior shape information. Starting from an initial
random state and an upper bound on the number of objects in the scene, the
optimization process eliminates redundant shapes by evolving them to empty
contours. The method operates on the probability maps produced by dense
semantic segmentation, taking advantage of object appearance prior information
learned from training examples. We instantiate the framework specifically for
fallen tree stem detection, evolving rectangular shapes according to the energy
functional which combines a nonparametric shape prior, a data fit term, and a
collinearity model. We propose a simulated annealing scheme with stochastic
sampling as the method of choice for evolving the optimal shapes and their
spatial orientations. The evaluated energy is defined on the space of
\emph{polygons}. The target polygons are obtained by finding contours of
0.5-superlevel sets of probability images from semantic segmentation.
This enables efficient computation of energy changes from applying neighboring
moves, because calculating intersections between the rectangular shapes and the
target polygons can be carried out with log-linear time complexity with respect
to the number of edges in the polygons~\citep{ZALIK2000137}, as opposed to
being a function of the number of image pixels.

The rest of this paper is organized as follows. In Section~\ref{sec:related}, we
report related work regarding both the detection of fallen trees from imagery
and methodological aspects of combining active contour methods with
CNN based segmentation. Section~\ref{sec:framework}
introduces the general framework for instance segmentation of images based on
multi-contour evolution on an abstract level, whereas in
Section~\ref{sec:specificMethod}, the framework is instantiated for detecting
fallen trees; we provide details of the tailored solution neighborhood operator
for the stochastic optimization, the initialization strategy as well as specific
realizations of the shape prior and other elements of the energy functional. In
Section~\ref{sec:experiment}, experimental evaluations of the proposed method
are provided, in comparison to a baseline operating on line level.
Also in this section we investigate the impact of using a CNN for generating the
appearance prior versus a simple baseline derived from raw image channel
intensities. The experimental results are discussed in
Section~\ref{sec:discussion}, and the most important conclusions are summarized in the final section.

\section{Related work}\label{sec:related}

To the best of our knowledge, this is the first contribution addressing the
large scale detection of fallen stems from aerial imagery on a polygon level,
which provides a comprehensive evaluation on over 700 reference polygons.
From an application standpoint, the two approaches conceptually most similar to
ours use the Hough transform to fit lines representing individual stems
in binarized images of target class posterior probabilities obtained on the
basis of hand-crafted textural features~\citep{rs9040306} or spectral
thresholding~\citep{Panagiotidis2019}.~\citet{rs12203293}
performed generic line detection within RGB orthomosaics
derived from very-high resolution unmanned aerial
system-acquired imagery to find approximate fallen stem shapes.\citet{f10060471}
used a generic segmentation procedure on the spectral bands of the aerial image combined with the normalized difference vegetation index~\citep{Tucker79}, and subsequently classified the resulting clusters based on spectral/textural features augmented with LiDAR derived information (canopy height model). \citet{f8010021} applied a
similar approach, using large-scale mean shift in the role of the segmentation
algorithm and augmenting the set of spectral bands with linear transformations
of raw bands, textural features and multiple vegetation indices. However,
neither of these approaches restricts the generic image segmentation to follow
the shape or appearance of fallen stems, therefore in case of multiple
intersecting trees, individual stems would not be delineated. We believe that a
key advantage of our proposed method versus generic segmentation approaches is
that the former has knowledge of the target object's shape, whereas the latter
do not. Therefore, while it may be possible to find parameters of generic
methods that produce acceptable segmentations for any particular scene, these
parameters (e.g. bandwidths, number of clusters) and not readily
learnable from training data or easily transferable between scenes. In contrast,
our method is informed on the dimensions of fallen stems as well as on the
interactions between them, allowing it to decompose the scene into objects which
plausibly look like fallen stems. On the area level, ~\citet{Latifi2018} used
synthetic RapidEye images to assess the extent of damage in spruce stands resulting from a bark beetle infestation.
Regarding the use of deep neural networks for detecting diseased and dead
trees, \citet{rs11060643} applied a CNN to classify tree vitality from patches
of RGB aerial imagery. \citet{s19071579} used the Faster R-CNN~\citep{7485869}
to detect regions of close-range images containing tree stumps, which were then
classified with respect to their root and butt-rot status.

On a more abstract level, our method could be interpreted as a way of
integrating CNNs with (multiple) active contour segmentation. Other ways of
achieving this were previously reported by several authors. In the
context of individual building segmentation from aerial imagery,
\citet{Marcos_2018_CVPR} proposed an end-to-end trainable framework utilizing
CNNs for learning the geometric prior parametrizations of an active contour
model (ACM). Inference from the ACM was integrated into the CNN weight update
schedule through computing a structured loss on the predicted and ACM's predicted output versus
ground truth polygons, and backpropagating the loss to the CNN parameters.

Our proposed approach borrows some ideas from the work of~\citet{Cremers2007},
where the active contour energy functional was designed to interact with the
input image indirectly through the intensity priors. The authors also directly
modeled the prior distribution of the shape coefficients using a kernel density
estimator. Our energy formulation shares some similarities with the energy
function utilized by~\citet{6520846} for multiple object tracking, which also
included a data fit term, pairwise interaction terms between tracked objects as
well as unary potentials encoding physical motion constraints (analogous to
proir information).

\section{Multi-contour segmentation with priors}\label{sec:framework}

We consider the generic problem of fitting multiple instances of a single object
class from abstract 'images'. Although all objects are by construction of the
same class, a reasonable amount of intra-class variation in shape as
well as appearance is allowed and expected. Usually, the image space
$\mathbb{I}$ will correspond to either the image plane $\mathbb{R}^2$ or 3D
Euclidean space $\mathbb{R}^3$, allowing to model e.g.~2D rasters or
(voxelized) 3D point clouds. However, any vector space is viable where there is
a meaningful concept of shape, rigid transformations (isometries) and a way of
measuring shape overlap. Denoting an input image as $I \in \mathbb{I}$ sampled
from the image space, we assume that $I$ contains an unknown number $M$ of
object instances from the target class $C$. Our goal is to retrieve
approximations of the target objects with respect to a pre-specified shape model
$P_s(\bar{\alpha})$ and its associated shape generator function
$f_s(\bar{\alpha})$, parameterized by a vector of abstract shape coefficients
$\bar{\alpha}$. The shape generator instantiates shapes in standard position
(centroid at the coordinate system origin, no rotations around axes). This function may be as
complex as a generative adversarial network, where the shape coefficients
represent the randomly sampled noise input, or as simple as a rectangle
generator parameterized by a width and a height. Additionally, each modeled
shape is equipped with its own set of position/orientation parameters $\theta_i$
which would typically include translations with respect to each coordinate axis
and appropriate rotations as required by the dimensionality of $\mathbb{I}$. We
will denote the shape generated by $f_s$ for coefficients $\bar{\alpha}$ and
rigidly transformed by $\theta$ as $f_s(\bar{\alpha}|\theta)$. 

In order to decouple shape and appearance (i.e.~image intensity) information, we
introduce an explicit discriminative prior $P_i(C|I)$ on the image space
$\mathbb{I}$. This image intensity prior transforms the original, possibly
multi-channel $I$ into a new \emph{probability image} $I_p$ encoding the class probabilities of
$C$ given the intensities. In practice, this can be seen as the output of a
semantic segmentation, like the U-net~\citep{RFB15a} in case of 2D raster
images, or VoxNet~\citep{7353481} for voxelized 3D point clouds. By extracting
contours of $q-$level supersets of $I_p$ (using e.g.~the marching cubes algorithm by~\citet{marchingCubes}), we may obtain a partition of $I$ into regions
corresponding to the target class, or 'foreground', versus 'background' regions. The comparison between shapes evolving according to the model $P_s$ and 'foreground' shapes
present within the image now boils down to the calculation of set intersections and
differences. Indeed, the shape model does not interact with the original image
$I$ other than through the extracted level supersets from $I_p$. 
We define the collection of connected regions inside
the probability image $I_p$ as $S=\{s_i \subset I_p, i=1\ldots n_s\}$,
corresponding to the extracted level supersets. The elements of $S$ form
polytopes of appropriate dimension, e.g.~polygons in 2D and polyhedrons in 3D.
Note that these polytopes need not represent single instances of the target
objects. In highly complicated scenarios, we expect them to consist of many
intersecting and overlapping instances.

\begin{figure*}[h]
\centering
\includegraphics[width=0.85\textwidth]{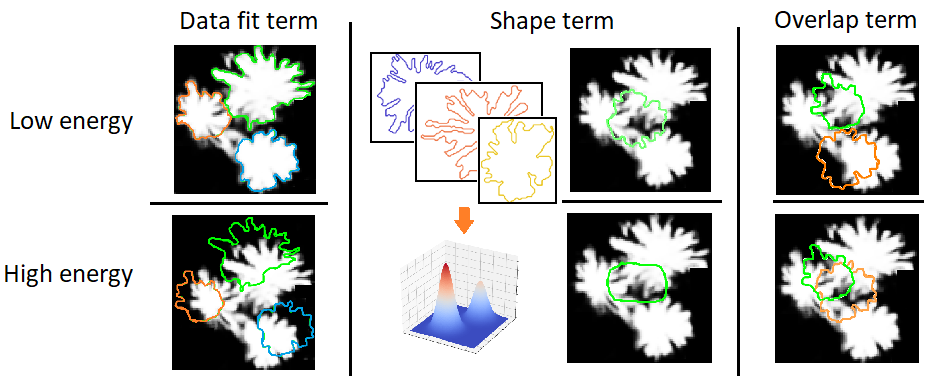}
\caption{Illustration of the impact of the various terms on the aggregate
energy function, which is a linear combination of the data fit, shape, and
overlap potentials. Left column: data fit potential ensures that a large
percentage of high-probability target class areas are covered by the evolving
contours. Middle column: shape potential ensures that the evolved shapes
are within the expected variability of the target objects' shape distribution.
Right column: overlap term prevents covering the same parts of the image with
different evolving contours.}
\label{fig:energyTerms}
\end{figure*}

\subsection{Energy function}\label{sec:energy}

Based on the definitions from the previous section, we are now ready to
introduce the energy function which drives the evolution of the modeled shapes.
Let $M'$ denote an initial overestimation of the true number of objects $M$
inside the input image. Then, each evolving shape is described by its vector of
shape coefficients $\bar{\alpha}_i$ as well as the location/orientation
parameters $\theta_i$. Collecting all models parameters into a vector
$\Omega=(\omega_i=(\bar{\alpha}_i, \theta_i))_{i=1\ldots M'}$, let $F(\omega_i)$
be an alias for $f_s(\alpha_i|\theta_i)$. The aggregate energy of the shape set
is given by Eq.~\ref{eq:totalEnergy}:
\begin{equation}\label{eq:totalEnergy}
\begin{split}
E(\Theta|S)=& \underbrace{\gamma_d E_{d} \left [\bigcup_{s \in S}s,
\bigcup_{i} F(\omega_i) \right ]}_{\text{data fit term}} -
\underbrace{\gamma_s \sum_{i} \log P_s(\bar{\alpha_i})}_{\text{shape
probability term}}\\
&+\underbrace{\gamma_o \sum_{j,k,j \neq k} E_o[F(\omega_j),
F(\omega_k)]}_{\text{pairwise overlap term}}\\
&+\underbrace{\sum_u \tau_u E_{aux,u}(\Omega)}_{\text{auxiliary potentials}} 
\end{split}
\end{equation}

An illustration of each energy term/potential's role and impact on the aggregate
energy is given by Fig.~\ref{fig:energyTerms}.

\subsubsection{Data fit potential}

The role of the data fit term $E_d(\tau, \phi)$ is to ensure that the model
shapes coincide well with the target class regions of the image. It is a
function of two sets:
(i) the union $\tau$ of all target class regions $s \in S$ extracted from the probability
image $I_p$, and (ii) the union $\phi$ of all currently modeled shapes obtained from 'decoding'
the elements of $\Omega$ with the generator function $f_s$ and applying the
respective rigid transform. Ideally, the sets (i) and (ii) should coincide,
however in practice the differences $\tau \setminus \phi$ as well as $\phi
\setminus \tau$ are non-empty. The former corresponds the
parts of regions designated as 'target class' that are not covered by any model
shape (false negatives). Symmetrically, $\phi \setminus \tau$ indicates regions
deemed as 'target class' by the model, but not intersecting with any elements $s \in S$, and thus
lacking evidence in the input image (false positives). The value of $\tau
\setminus \phi$ impacts the specificity/recall of the segmentation,
whereas  $\phi \setminus \tau$ impacts the sensitivity/precision. We allow an
assignment of different weights to these two quantities, reflecting the fact
that the tradeoff between precision and recall may be asymmetric for some
applications:
\begin{equation}\label{eq:dataFitPotential}
E_d(\tau, \phi) = 2[(1-\pi_p)\lambda(\tau \setminus \phi) + \pi_p \lambda(\phi
\setminus \tau)]
\end{equation}
In the above, the term $\lambda(\cdot)$ can be thought of as analogous to the
Lebesgue measure on the Euclidean space of the appropriate dimension, i.e. area
in 2D, volume in 3D, etc. The term related to the precision (false positives) is
weighted with $0 \leq \pi_p \leq 1$.

\subsubsection{Shape probability potential}

This term is directly derived from the prior shape model $P_s(\bar{\alpha})$ as
the sum of negative log-likelihoods of all model shapes. It acts as a
regularizer for the shape coefficients, penalizing shapes which become too
unlikely with respect to the learned prior. Note that for some generator
functions, the shape coefficients $\bar{\alpha}$ may already be distributed
uniformly \emph{by construction} inside the (appropriately scaled) unit
hypercube, in which case the shape probability term boils down to a constant and
may be removed. For an example, see
e.g.\citep{isprs-archives-XLIII-B2-2020-717-2020}, where a generative
adversarial network with uniformly distributed latent variables was used as the shape model within the active contour segmentation framework.

\subsubsection{Overlap potential}

Since our framework assumes that initial guess on the number of shapes $M'$ is
biased towards too high values, we expect that part of the model shapes will
become redundant. To prevent duplicate coverage of the same image regions by
different model objects, and to allow the number of active shapes to converge to
the true number of objects present within the image, we define an overlap
potential $E_o$ which penalizes overlapping of model shapes. Acting together
with the data fit term $E_d$, it is designed to direct a redundant shape
towards evolving into an empty contour: $E_o$ will push a shape away from areas
already occupied by other model shapes, while $E_d$ will ensure that the shape
does not occupy background regions of the input image. Note that not all forms
of overlap should be penalized. For example, in our application of fallen tree
segmentation, intersections of tree stems that are not parallel to each other
are unlikely to be due to instance duplication, instead they are the result of
physical overlap and stacking. To model this, we utilize an auxiliary term
$\kappa(o_1, o_2) \in [0;1]$ which quantifies the likelihood of model shapes
$o_1,o_2$ belonging to the same real-world object. The overlap potential
is then defined in a pairwise manner as:
\begin{equation}\label{eq:overlapPotential}
E_o(o_1, o_2) = \kappa(o_1, o_2) \lambda(o_1 \cap o_2)
\end{equation}
The potential $E_o$ is evaluated over all pairs $i,j \in {1,2,\ldots,M'}$ such
that $i < j$, with each value contributing to the total energy $E(\Theta)$ in
equal proportion. Once again, $\lambda$ indicates the 'natural' measure in the
Euclidean space of the appropriate dimension (area, volume etc.).

\subsubsection{Auxiliary potentials}

Our framework allows for application specific potentials $E_{aux,u}$, each
weighted by their own coefficient $\tau_u$. In our formulation, to maintain the
highest flexibility, these potentials are functions of the entire model
parameter vector $\Omega$, which means they have access to both the shape
coefficients and the decoded/transformed model shapes. This formulation admits
unary, pairwise, or even higher order potentials. In
Section~\ref{sec:specificMethod}, an example of an auxiliary pairwise potential
is shown, which is designed to discourage collinearity between the modeled
stems.

\subsection{Relationship to active contour segmentation}

The proposed framework can be viewed as one possible generalization of the
classic foreground-background active contour
raster image segmentation~\citep{CremersIJCV} to multiple object instances and
more general images. The statistical formulation by~\citet{Cremers2007} assumed that
the evolving contour of a target class $C$ region has an abstract shape
parametrization $\bar{\alpha}$, endowed with a prior model $P_s(\bar{\alpha})$.
The optimized energy functional represented a trade-off between the data fit
term and probability of the evolving shape. Denoting $H_{\bar{\alpha}}[x]$ as
the indicator function for image element $x$ lying inside the shape, their
energy objective can be written as:
\begin{equation}
\begin{split}
E'(\bar{\alpha}) =& - \int ( H_{\bar{\alpha}}[x] \log P(x|C)\\ 
&+ (1-H_{\bar{\alpha}}[x]) \log P(x|\neg C)) dx\\
 &- \log P_s(\bar{\alpha})
\end{split}
\end{equation}

Here, the foreground and background regions have their separate image intensity
likelihoods $P(x|C), P(x|\neg C)$. By assuming equal prior probabilities on the
foreground/background regions ($P(C) = P(\neg C)$), based on the Bayesian rule
we can express the data fit potential in terms of the target class
\emph{posterior} $P(C|x)$, resulting in:

\begin{equation}\label{eq:originalACM_stat}
\begin{split}
E''(\bar{\alpha}) &= - \underbrace{\int  H_{\bar{\alpha}}[x] \log P(C|x)
dx}_{\text{data fit inside contour}}\\
&- \underbrace{\int (1-H_{\bar{\alpha}}[x]) \log [1-P(C|x)]) dx}_{\text{data fit
outside contour}}\\
 &- \log P_s(\bar{\alpha})\\ 
 &= E'(\bar{\alpha}) + D
\end{split}
\end{equation}
The new energy function $E''$ differs from $E'$ only by a constant value $D$,
therefore their extrema coincide (see
e.g.~\citet{isprs-annals-II-3-W4-181-2015}). Moreover, under certain
assumptions, the data fit terms inside and outside of the contour are analogous
to the quantities $\lambda(\tau \setminus \phi)$ and $\lambda(\phi \setminus
\tau)$ from our data fit potential (Eq.\ref{eq:dataFitPotential}). Specifically, recall that the
target connected regions $s_i \in S$ are high-probability $q-$level supersets
extracted from the probability image. We can therefore view the posterior class
probability inside and outside these regions as respectively $p_{in} \approx
1-\epsilon, p_{out} \approx \epsilon$, where $\epsilon$ is a small positive
constant. In this setting, the intersection of all objects modeled by the
current state $\Omega$ with the set of target regions $S$ will contribute
$\lambda(\tau \cap \phi)\cdot \log (1-\epsilon) \approx 0$ to the data fit
potential, since $\log (1-\epsilon)$ tends to 0 with $\epsilon$. On the other
hand, the difference $\phi \setminus \tau$ will contribute $\lambda(\phi
\setminus \tau)\cdot \log \epsilon$, which tends to $-\infty$ as $\epsilon$
tends to zero. By a similar argument, one can show that the data fit term
outside the contour from Eq.~\ref{eq:originalACM_stat} is dominated by the
symmetrical expression $\lambda(\tau
\setminus \phi)\cdot \log \epsilon$. Setting $\pi_p$ to 0.5 in our data fit
potential (Eq.~\ref{eq:dataFitPotential}), we see that an instantiation of our
framework with a single modeled object is equivalent to the original statistical
active contour formulation with probabilities quantized at ${1-\epsilon,
\epsilon}$ such that $\gamma_d = -\log \epsilon$.

\subsection{Optimization}\label{sec:optimization}

To optimize the total energy from Eq.~\ref{eq:totalEnergy}, various stochastic
and combinatorial techniques are available based on the choice of quantization
or lack thereof for the model variables. If all shape and rigid transformation
parameters are continuous, Eq.~\ref{eq:totalEnergy} can be minimized using
stochastic methods like simulated
annealing~\citep{Kirkpatrick1983OptimizationBS} or a hybrid Monte Carlo-gradient
based approach like basin hopping~\citep{doi:10.1021/jp970984n,Li6611}. The
latter is particularly useful in settings where the gradients of all the energy function
terms with respect to all model variables may be computed analytically.
Sometimes it may be reasonable to discretize the domain of one or more
variables, e.g.~the shape translation parameters from $\theta_i$ could be
expressed in pixels or voxels. In such cases, generic metaheuristics for solving
mixed combinatorial/continuous problems are applicable, including methods from
the class of evolutionary algorithms, specialized versions of tabu search
(e.g.~\citep{Siarry1997}), and simulated annealing.

The aforementioned metaheuristics are based on exploring the neighborhood of the
current solution and making local moves which alter a small part of it. This
makes it cumbersome and inefficient to apply these local-search based methods to
the minimization of our energy (Eq.~\ref{eq:totalEnergy}), since the data fit
term requires a union of all of the evolving shapes $\phi=\bigcup_i
F(\omega_i)$, and an intersection of this union with the high-probability object contours from the
input image $\tau=\bigcup_{s \in S} s$. Even if only one shape $\omega_i$ is
altered, all of the aforementioned calculations need to be repeated to re-evaluate the
data fit potential $E_d$. To localize the effects of modifying individual shapes
and make local search steps more efficient, we propose the following
approximation. First, observe that as $\tau$ does not depend on the model
variables, it can be precomputed once and reused in the calculations. Moreover,
we may express the intersection $\phi \cap \tau$ as a union of intersections $\bigcup_i
\tau \cap F(\omega_i)$. Applying the inclusion-exclusion principle, we can
write:
\begin{equation}\label{eq:inclExcl}
\begin{split}
|\phi \cap \tau|&=\underbrace{\sum_i |\tau \cap F(\omega_i)|}_{\text{unary
term}} - \underbrace{\sum_{i \leq i < j} |\tau \cap F(\omega_i) \cap
F(\omega_j)|}_{\text{pairwise term}}\\
&+\underbrace{\sum_{k=3}^{M'}(-1)^{k+1}(\sum_{1 \leq i_1 < \ldots < i_k} |\tau
\cap F(\omega_{i_1})\cap \ldots \cap F(\omega_{i_k})|)}_{\text{residual}}
\end{split}
\end{equation}
We choose to approximate $|\phi \cap \tau|$ by its underestimation given by the
first two terms (unary and pairwise) in the inclusion-exclusion expansion
(Eq.~\ref{eq:inclExcl}). This corresponds to ignoring contributions from subsets
where 3 or more of the model shapes intersect. In practice, we believe this
approximation is sufficient, because the model explicitly discourages overlap of
multiple shapes through the overlap potential $E_o$. Moreover, by using an
underestimation of the true intersection area $|\phi \cap \tau|$, the
multi-shape overlap is penalized even more due to the subtraction of the
overlapping area in the pairwise term and not recovering it in the (removed)
residual. This leads the model away from undesirable overlap. However, the main
benefit is that the influence of changing a single model shape $i$ (by mutating
its shape coefficients or rigid transform parameters) is now reduced to
affecting one unary term $|\tau \cap F(\omega_i)|$ and at most $M'$ pairwise
terms $|\tau \cap F(\omega_i) \cap F(\omega_j)|, j \in {1,\ldots,M'}$. The pairs
$i,j$ which do not intersect can be filtered out using simple
bounding box criteria. Combined with caching the values $|\tau \cap
F(\omega_i)|, |\tau \cap F(\omega_i) \cap F(\omega_j)|$ for all model objects
and their pairs, the term $|\tau \setminus \phi|$ can be efficiently updated
based on the values of $|\tau \cap \phi|, |\tau|$, by using the identity
$|A \setminus B| = |A| - |A \cap B|$. In a similar manner, the value of $\phi$
may be approximated by caching and updating the first and second-order terms of the
inclusion-exclusion expansion $|F(\omega_i)|, |F(\omega_i) \cap F(\omega_j)|$,
yielding $|\phi \setminus \tau|$. Moreover, the caching of the pairwise terms
$|F(\omega_i) \cap F(\omega_j)|$ also enables fast updates of the term $E_o$.
Finally, the shape model $P_s$ is already in additive form, therefore altering 
$\bar{\alpha_i}$ influences only the $\log P_s(\bar{\alpha_i})$.
Care should be taken when instantiating auxiliary potentials to ensure that they also allow efficient partial updates,
leading to applicability of local solution perturbation based metaheuristics.

\begin{figure*}[h]
\centering
\includegraphics[width=0.95\textwidth]{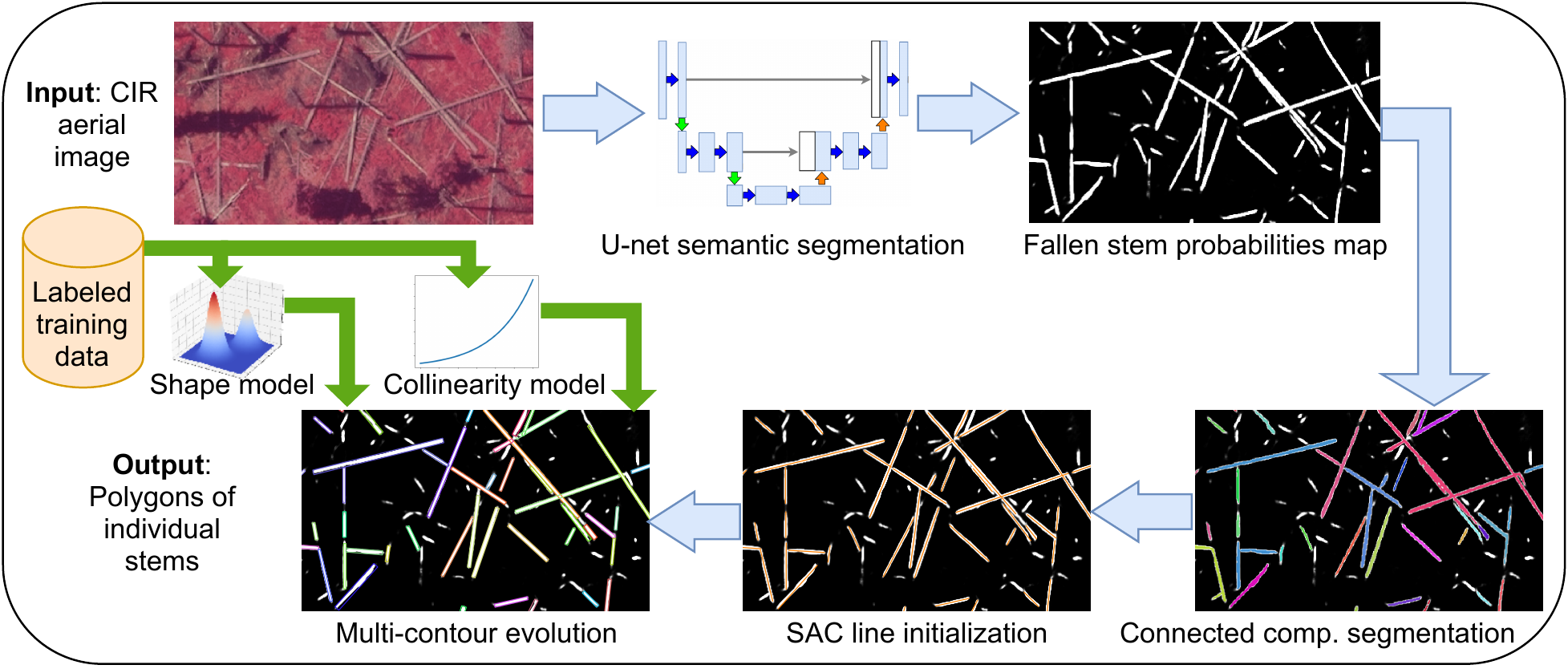}
\caption{Overview of the processing pipeline for delineating individual stem
polygons using multiple active contour evolution. The input CIR image undergoes
semantic segmentation using the U-net, and the fallen stem probability map is
partitioned into high-probability connected components. Next, the model shape
positions and dimensions are initialized based on sample consensus line
segmentation. Finally, the model shape configuration is optimized using
simulated annealing, under consideration of the shape and collinearity models
learned from labeled training data.}
\label{fig:strategyOverview}
\end{figure*}

\section{Application of our framework to the instance segmentation of fallen
trees}\label{sec:specificMethod}

In this section, we instantiate the framework described in the previous chapter,
obtaining a method for detecting individual fallen stems from aerial imagery. We
consider a 2D raster image with $N_{c}$ channels as input. Ideally, the image
should include a near-infrared channel, which is known to differentiate dead and
living vegetation well. The rest of this section explains the
instantiation of various components of the energy term, the strategy for
exploring the solution space using the neighborhood operator, as well as an
initialization scheme based on detecting lines using the sample consensus
method. An overview of the entire processing pipeline is depicted in
Fig.~\ref{fig:strategyOverview}.

\subsection{Fully convolutional networks}\label{sec:fcn}

\begin{figure}[h]
\centering
\includegraphics[width=1.0\columnwidth]{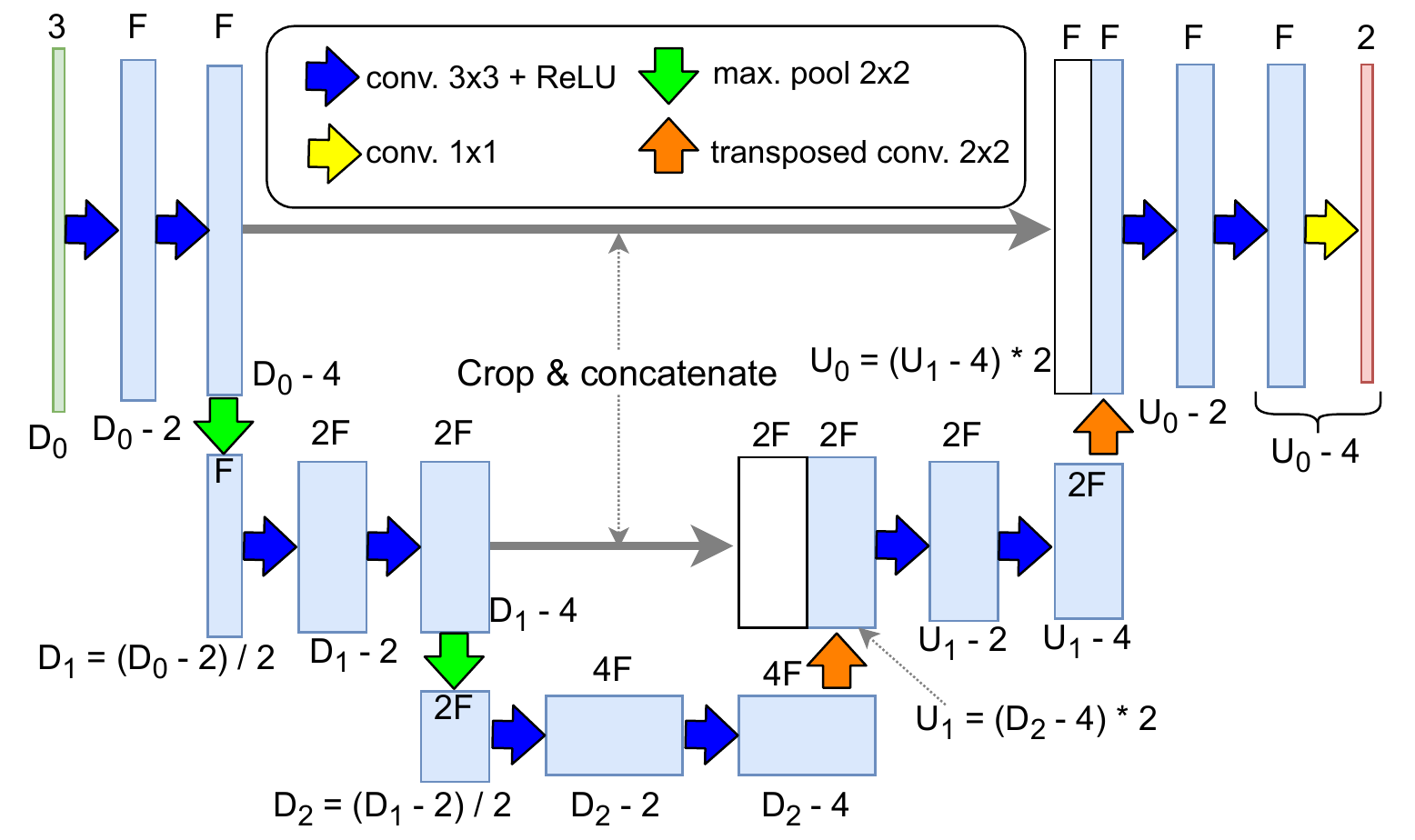}
\caption{Architecture of a 3 layer U-net for binary classification of 3-channel
images. At level $k$, the layers undergo convolution with a series of 3x3
filters, producing $2^kF$ feature maps. The initial size $D_0$ of the input
image is approximately halved at each downsampling layer, an approximately
doubled in each upsampling layer (up to border removing convolutions). The
final layer is obtained by a 1x1 convolution with the top-level upsampled
feature layer, and is subsequently fed into the softmax operator to derive
class posterior probabilities.}
\label{fig:unet}
\end{figure}

Fully convolutional networks (FCNs) are a class of convolutional
artificial neural networks (CNNs) designed for dense semantic segmentation of
raster images. As opposed to classical CNNs that are used primarily for image (sparse) classification
(i.e.~assigning a single label to an entire image or patch), FCNs do not possess
fully connected layers, which makes them independent of the input image
size~\citep{long2015fully}. FCNs primarily consist of convolutional/transposed
convolutional filters as well as pooling layers, organized into two symmetrical
paths. The encoder path downsamples the original image into meaningful
features by means of convolutional filters and pooling operations,
whereas the upsampling path aims at decoding these features into a full-sized
output map using transposed convolution operations. The final, topmost
upsampling layer of the network is fed into a softmax operator, producing
per-class posterior probabilities at each pixel and enabling end-to-end training
with a logistic loss function. A classic FCN architecture which attained widespread use
across various applications~\citep{akeret2017radio,10.1007/978-3-319-60964-5_44}
is the U-net~\citep{RFB15a}, where upsampling layers are augmented with feature
maps from the downsampling path at the corresponding resolution,
to provide more context information. The architecture of a classic U-net is
depicted in Fig.~\ref{fig:unet}. It should be noted that due to the handling of
image borders in the convolution operation, a decrease in image size occurs at
each convolutional filtering layer in both the downsampling and encoding
branches. This results in the output network layer having smaller dimensions
than the original input image. To process input images of arbitrary size, a
tiling strategy must therefore be employed, where input windows for subsequent
applications of the U-net overlap by the margin derived from the difference
between input and output layer shapes (see \citet{RFB15a} for details).

\subsection{Image intensity prior}\label{sec:intPrior}

In the role of the image intensity prior $P_i(C|I)$ for our target class of
fallen trees, we utilize the U-net deep neural network~\ref{sec:fcn}
in a binary classification setting. The original architecture is easily
adaptable to the variable number of input channels $N_{c}$. The
per-pixel posterior object class probabilities conditioned on the image pixel
intensities (i.e.$P_i(C|I)$) are obtained directly from the semantic
segmentation.
We subsequently apply the marching squares algorithm~\citep{marchingCubes} to
derive contours of $q-$level supersets of the probability image. This results in
a set of high-probability polygons, possibly consisting of multiple fallen trees and non-class objects or noise.
Since the best known geometric algorithms used for calculating polygon
intersections have a worst-case computational complexity proportional to the 
product of their vertex counts in the general (non-convex)
case~\citep{10.1145/358656.358681}, we apply the contour simplification
algorithm by~\citet{PEUCKER73} (parameterized by the max.~simplification
distance $\epsilon_d$) to the polygons, resulting in the final set $S$ defined
in Sec.~\ref{sec:framework}.

\subsection{Shape generator}

As the shapes of fallen stems are well approximated by
rectangles, we utilize a simple shape generator
$f_s(\bar{\alpha})$ parameterized by two scalars $\alpha=[a,b]$, which produces
a rectangle with side lengths $a,b$ centered at the origin of the coordinate
system, oriented parallel to its axes (i.e.~in standard position):
\begin{equation}
\begin{split}
&f_s(a,b) = [(-\frac{a}{2};-\frac{b}{2}), (\frac{a}{2};-\frac{b}{2}),
(\frac{a}{2};\frac{b}{2}), (-\frac{a}{2};\frac{b}{2})]\\
&f_s(a,b;\theta=[x_c,y_c,\rho]) = R_\rho f_s(a,b) + T_{x_c, y_c}
\end{split}
\end{equation}

The rigid transformation parameters $\theta$ consist of the
center position $(x_c, y_c)$ translation $T$ and an in-plane rotation $R$ by
angle $\rho$.

\begin{figure}[h]
\centering
\includegraphics[width=0.6\columnwidth]{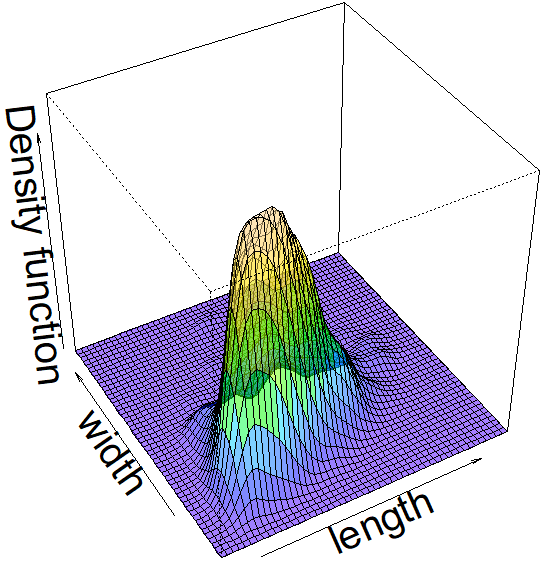}
\caption{Sample kernel density estimator model of joint stem length/width
probability based on reference labeled polygons.}
\label{fig:shapeModelVis}
\end{figure}

\subsection{Energy components}\label{sec:energyComponents}

Here we provide details about component potentials of the energy function
(Eq.~\ref{eq:totalEnergy}). Aside from the 3 standard potentials defined in
Section~\ref{sec:framework}, we introduce an auxiliary collinearity potential to
help prevent the fragmentation of object detections into multiple collinear
parts.

\subsubsection{Data fit and overlap terms}

We utilize the aforementioned (Sec.~\ref{sec:optimization}) second-order
inclusion-exclusion principle based formulation to approximate the set
difference cardinalities $\phi \setminus \tau, \tau \setminus \phi$ by
appropriate pairwise intersections. Since the model polygons have a constant
dimension of 4 vertices, the computation of any pairwise intersection of model
polygons $i$ and $j$, $f_s(\alpha_i|\theta_i) \cap f_s(\alpha_j|\theta_j)$ may be done in
constant time, whereas the computational complexity of intersecting any
$f_s(\alpha_i|\theta_i)$ with a high-probability contour $s_i \in S$ is linear
in the number of vertices forming $s_i$~\citep{10.1145/358656.358681}.
Additionally, we modify the generic overlap potential $E_o$
(Eq.~\ref{eq:overlapPotential}) to include a dependency on the angular
difference in orientations between the model shapes:
\begin{equation}
E_o(i,j) =
e^{\frac{-(\rho_i-\rho_j)^2}{2\sigma_o^2}} |f_s(\alpha_i|\theta_i) \cap
f_s(\alpha_j|\theta_j)|
\end{equation}
This reflects the model's capability to allow non-parallel, crossing
shapes to overlap without being penalized, as they most likely do not
correspond to the same object (fallen stem).

\subsubsection{Shape prior}\label{sec:shapePriorSpecific}

We utilize a shape prior model in the form of a kernel density estimator defined 
on the shape coefficients $\bar{\alpha}=[a,b]$, based on a set of training rectangle shapes
$S_T=\{\bar{\alpha}_k\}$:
\begin{equation}
\label{eq:kdeWt}
P_s(\bar{\alpha})=\frac{|H|^{-1/2}}{|S_T|}\sum_{k=1}^n K(H^{-1/2}(\bar{\alpha} -
\bar{\alpha}_k))
\end{equation}
In the above, the bivariate Gaussian kernel is applied in the role of
$K$, whereas the bandwidth matrix $H$ is determined via the plug-in selection
method of~\citet{WandJones1994}. A sample shape model derived from part of our
reference shapes is depicted in Fig.~\ref{fig:shapeModelVis}.

\subsubsection{Collinearity prior (auxiliary potential)}\label{sec:collin}

\begin{figure}[h!]
\centering
\begin{subfigure}[b]{1.0\columnwidth}
	\includegraphics[width=1.0\columnwidth]{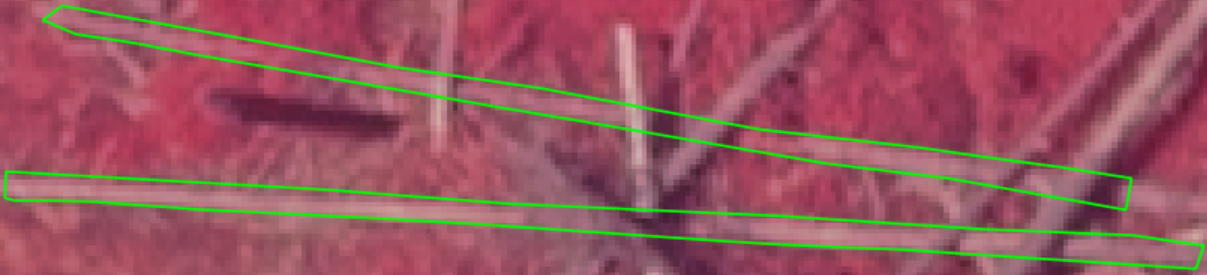}
	\caption{}
\end{subfigure}
\begin{subfigure}[b]{1.0\columnwidth}
\centering
\includegraphics[width=1.0\columnwidth]{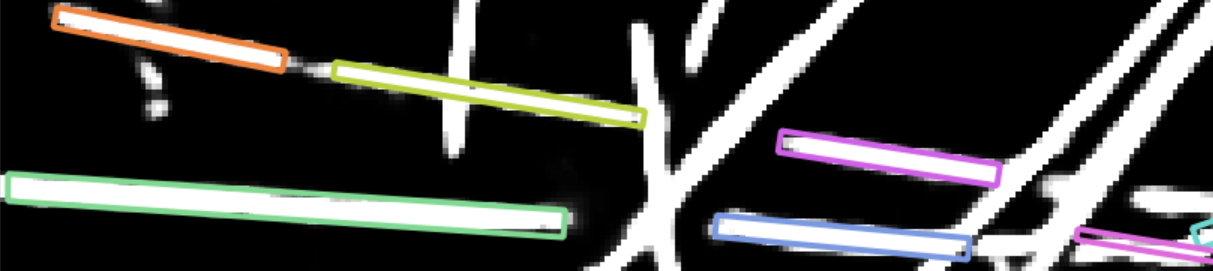}
	\caption{}
\end{subfigure}
\caption{(a) Color infrared image of forest scene with fallen stems. Two long
stems are marked with green outlines. (b) Sample detection result for the two
stems over posterior class probability image of same scene. Due to occlusions,
the stems are fragmented and discontinuous within the probability map, which
causes the energy function to prefer multiple disconnected collinear fragments
over a single polygon covering the full length of the stem.}
\label{fig:fragmentedStems}
\end{figure}

While the overlap penalty will discourage the formation of highly overlapping
model shapes, there is still a possibility of segmenting a single tree stem as a
sequence of nearly-collinear parts (see Fig.~\ref{fig:fragmentedStems}). To
mitigate this, we introduce an auxiliary pairwise collinearity potential
$E_c=\sum_{i,j} E_c(i,j)$, which penalizes highly collinear shapes located in
close proximity with each other.
Here, we define $E_c(i,j)=\log P_{eq}(f_s(\alpha_i|\theta_i),
f_s(\alpha_j|\theta_j))$ as the log-probability $P_{eq}$ of the two shapes $i,j$
belonging to the same stem. In practice, we use the output of a probabilistic
classifier (e.g.~logistic regression) acting on differential features derived
from the shapes' locations (angular deviation of orientations, mean average
distance of central axes). This is in analogy to the object similarity function
applied to graph cut segmentation of stem parts into individual fallen trees
defined in our prior work~\citep{Polewski2015b}. However, the log-probability
contributes positive values to the energy for each detected collinear shape
pair, thereby biasing the model away from such states and encouraging a merge operation of
the interacting shapes.

\subsection{Initialization with sample consensus}\label{sec:sampleConsInit}

\begin{figure}[h]
\centering
\includegraphics[width=1.0\columnwidth]{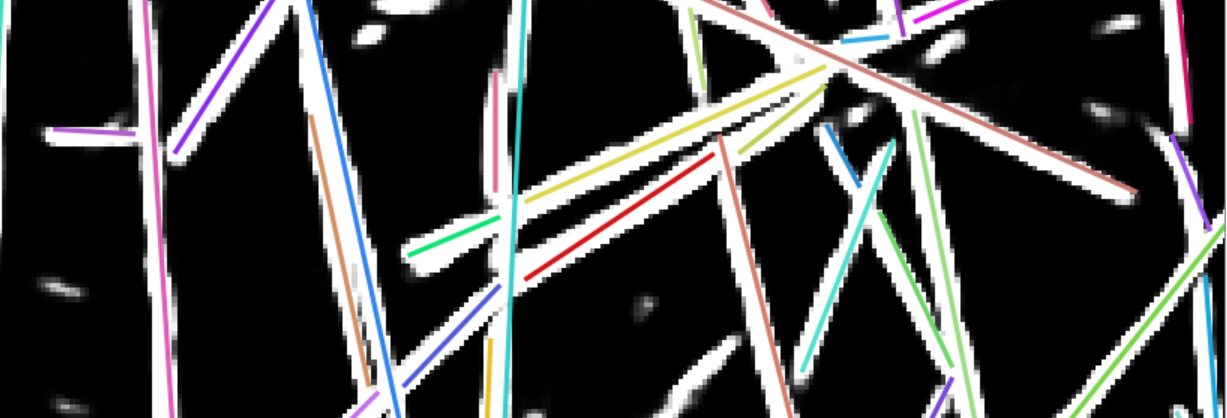}
\caption{Line segments discovered using an iterative sample consensus (SAC)
method. Although most target class pixels are covered by at least one SAC line,
the method usually overestimates the true number of stems due to the
variability of stem width distributions and the resulting difficulty in
defining a single inlier threshold appropriate for all cases.}
\label{fig:sampleSAClines}
\end{figure}

We initialize our model with a set of line segments automatically detected using
sample consensus (SAC) methods~\citep{10.1145/358669.358692} within the
binarized probability image $P_i(C|I)$ obtained from semantic segmentation (see
Sec.~\ref{sec:intPrior}). The inlier threshold for $d_{sac}$ SAC is set to the
maximum expected width of a fallen stem (expressed in pixels). We only allow
line segment hypotheses having a minimal length $l_{sac}$, again derived from
the minimal length of a tree stem we expect to find. This segment length is
measured as the length of the interval of inlier pixel projections onto the
respective model line. We also impose a minimum
number of inlier points $n_{sac}$ for valid hypotheses. The whole scene is
processed iteratively, greedily picking the highest-inlier hypothesis until
there are no valid hypotheses left. We expect to discover an overabundance of
line segment hypotheses, partially covering the vast majority of true stem
segments within the scene (see Fig.~\ref{fig:sampleSAClines}). Each accepted
SAC hypothesis becomes an initial model shape, with the length $l_0^i$ and
position $t_{x,0}^i, t_{y,0}^i$ / orientation $\rho_0^i$ inherited from the SAC
line and a default assigned width.
It is up to our energy formulation to eliminate redundant model elements,
determine true dimensions of each object, and improve delineation of individual
stem boundaries.

\subsection{Efficient evaluation of neighboring solutions}

\begin{figure}[h]
\centering
\includegraphics[width=1.0\columnwidth]{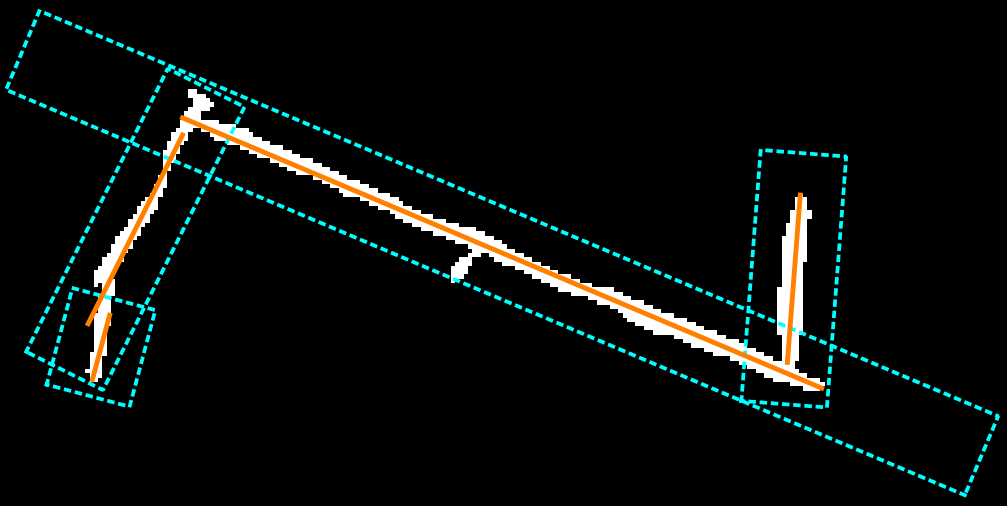}
\caption{Probability image with detected initial sample consensus hypotheses
(orange lines). Each line is surrounded by its center constraints polygon
(cyan boxes). The evolution of the model shape associated with the given
hypothesis line is constrained to maintain the center of the shape within the
corresponding box at all times.}
\label{fig:centerBounds}
\end{figure}

As our fallen tree detection pipeline is designed for processing high-resolution
nadir-view aerial imagery, it seems reasonable to quantize some size and
position parameters at the ground sampling distance (GSD) of the input image,
i.e.~the finest level of detail available within the image. Given the 
elongated shape of our target objects (fallen tree stems), this quantization can
yield a significant reduction of some parameters' domains. Specifically,
assuming a typical GSD of high-resolution aerial imagery of 5-10 cm, and the
diameter of fallen stems bounded by 70 cm, the width parameter of the generated
rectangle would only admit a small number (7-14) of possible values. For this
reason, we restrict the elements $a,b$ of the shape coefficient vector
$\bar{\alpha}$ and to the integer domain,
representing the number of pixels at the original image resolution. The
rectangle length and width $a,b$ are additionally equipped with their own
lower/upper bounds $[a_{lo};a_{hi}],[b_{lo};b_{hi}]$ based on the image
resolution and the expected maximum/minimum stem dimensions. The lower bound
$b_{lo}$ of the width is set to zero, which allows the model to 'disable' a
particular, redundant shape altogether and prevent it from contributing to the
energy function (through zero overlap with any other polygons). Additionally, we
impose restrictions on the location of the centers $t_{x}^i,t_{y}^i$ for every
shape $i$ separately, based on the centers of their SAC line segment
initializations $t_{x,0}^i, t_{y,0}^i$ (see previous section). The center of the
model shape $t_{x}^i,t_{y}^i$ must be located within a rectangle of length
$l_0^i$ oriented according to the initial SAC angle $\rho_0^i$ and having a
width $w_0$ which is a parameter of our method (see
Fig.~\ref{fig:centerBounds}).
We introduce these constraints in order to avoid drifting away from the initial
SAC solutions into poor regions of the solution space where no overlap of model
objects with high probability level-supersets of the image would exist and hence
loss of gradient would occur. The optimization method of choice, simulated
annealing, is susceptible to this kind of behavior during the initial phases of
the minimization process, where the temperature parameter remains high and even
poor moves which deteriorate the solution quality continue to be accepted.

\subsubsection{Solution altering moves}

Consider the state $\Omega^k$ of the solution at iteration $k$ of the
optimization process, consisting of all shape and rigid transformation
parameters concatenated into a single vector:
$\Omega^k=(\omega_i^k=(\bar{\alpha}_i^k, \theta_i^k))$ (see
Sec.~\ref{sec:energy}). To generate a new candidate state $\Omega^{k+1}$ from
$\Omega^{k}$, we designed the following solution altering moves, acting on a
random shape $\omega_u^k$:

\begin{enumerate}[i]
  \item length/width: add/subtract a random integer bounded by
  $\delta_l,\delta_w$ respectively to the length/width of shape $u$ 
  \item angle: add/subtract a random real number bounded by $\delta\rho$
  to the angle $\rho_u$ related to shape $u$'s orientation
  \item location (along axis): shift the center of shape $u$
  \emph{along its current axis} by a random number bounded by $\delta_{t,ax}$
  \item location (arbitrary): shift the center of shape $u$ by an arbitrary
  random 2D vector, the components of which are bounded by $\delta_x, \delta_y$ 
  \item merge/absorb: for a collinear shape $\omega_v^k$, extend the shape $u$
  by projecting the vertices of both shapes onto the current axis of shape $u$
  and adjusting its length and center point such that $u$ contains all the
  projections. Also, disable the contributions of shape $v$ to the overall
  energy by setting its width to zero
\end{enumerate}

The selection of the move to apply is based on a uniform random choice, where
the merge/absorb move is only considered if the probability $P_{eq}$ of two shapes belonging to
the same object is above a threshold value (see Sec.~\ref{sec:collin}). Moves
(i)-(iv) are of a local nature in the sense that only the model shape
$u$ changes. To calculate the new energy, we only need to perform a series of
constant-time rectangle intersection computations between $u$ and the remaining
shapes, as well as one or more intersection calculations between $u$ and the
high-probability object contours $s \in S$, linear in the respective vertex
counts. The values of the remaining model shape and image contour intersections
remain unchanged and can be cached as described in Sec.~\ref{sec:optimization}.
In case of move type (v), a similar technique can be applied, because while
technically two shapes are altered, only the 'absorbing' shape $u$ needs to have
its intersections recalculated since shape $v$ becomes an empty contour whose
intersection with an arbitrary polygon yields the empty set.

\begin{figure*}[ht]
  \centering
  \begin{subfigure}[b]{0.43\textwidth}
  	\includegraphics[width=\textwidth]{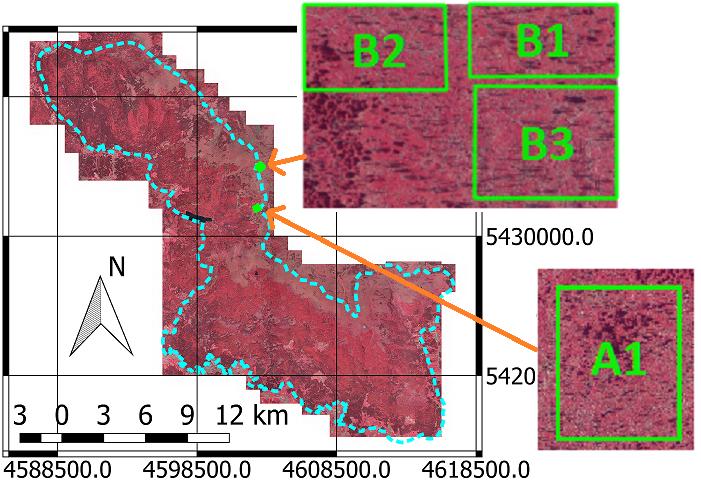}
  	\caption{}
  	\label{fig:natParkOverview}
  \end{subfigure}
  \hspace{\fill}
  \begin{subfigure}[b]{0.55\textwidth}
  	\includegraphics[width=\textwidth]{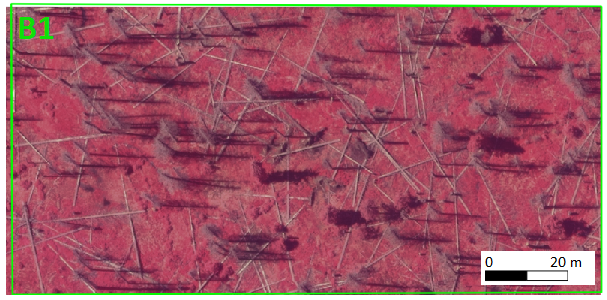}
  	\caption{}
  	\label{fig:testAreaB1}
  \end{subfigure}
  \caption{(a) Training and validation regions (in green) chosen within the
  Bavarian Forest National Park (boundaries shown with dashed cyan line). The
  coordinates and true north arrow are with respect to the coordinate reference
  system DHDN/3-degree Gauss-Kr{\"u}ger zone 4 (EPSG:31468). Background is color
  infrared image with ground sampling distance of 10 cm.
  Region A was used exclusively for training, whereas region B was the basis
  for validation. (b) a depiction of test area B1, containing a mixture of
  lying dead trees, standing dead trees, and living vegetation.}
  \label{fig:natParkAreas}
\end{figure*}

\section{Experiments and results}\label{sec:experiment}

In this chapter, we describe the source imagery, target training and test areas,
reference data, evaluation strategies, and the details of our experimental
setup used to evaluate the proposed dead tree delineation framework against a
baseline method. We also list the principal numerical results.

\subsection{Data acquisition}

For validating our method, we utilized aerial imagery from the
Bavarian Forest National Park, situated in South-Eastern Germany ($49^\circ 3' 19''$
N, $13^\circ 12' 9''$ E). The Bavarian
Forest lies in the mountain mixed forests zone consisting mostly of Norway
spruce (\emph{Picea abies}) and European beech (\emph{Fagus sylvatica}). From
1988 to 2010, a total of 5800 ha of the Norway spruce stands died off because
of a bark beetle (\emph{Ips typographus}) infestation~\citep{Lausch2012}. Color infrared images were acquired in the
leaf-on state during a flight campaign carried out in June 2017 using a Leica
DMC III high resolution digital aerial camera with a nadir across track field of view
of 77.3$^\circ$ (see~\citet{LeicaDMCIII} for the product sheet). Multiple
multispectral color cameras were utilized to form composite images, which had a
resolution of 14592 x 25728 pixels with a virtual pixel size of 3.9 $\mu m$ on
the CMOS sensor.
The mean above-ground flight height was ca.~2879 m, resulting in a pixel
resolution of 10 cm on the ground. The flight campaign took place between
10:30 and 13:25, with the sun's position traversing the range
49$^\circ$64$^\circ$35$^\circ$. The images contain 3 spectral bands: near
infrared (spectral range 808-882 nm), red (619-651 nm) and green (525-585 nm).
All digital CIR images were radiometrically corrected by using optimal camera
calibration observations, transformation parameters and ground control points.
The procedures were conducted in the program system OrthoBox (Orthovista,
Orthomaster) of the company Trimble/INPHO.

\begin{figure}[h!]
\centering
\begin{subfigure}[b]{1.0\columnwidth}
	\includegraphics[width=1.0\columnwidth]{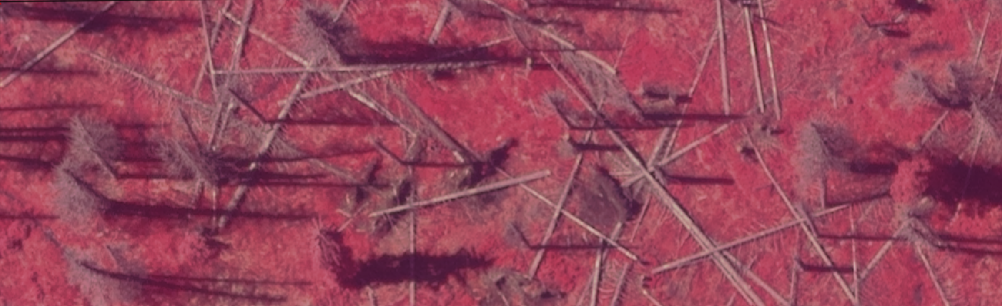}
	\caption{}
	\label{fig:labelArea}
\end{subfigure}
\begin{subfigure}[b]{1.0\columnwidth}
\centering
\includegraphics[width=1.0\columnwidth]{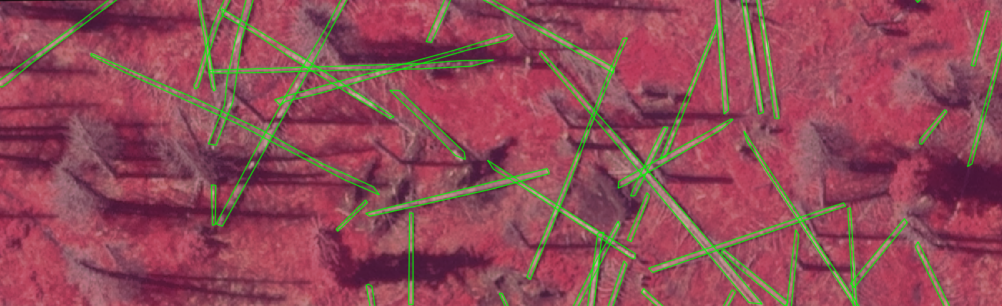}
	\caption{}
	\label{fig:labelPolyArea}
\end{subfigure}
\caption{(a) Sample color infrared (CIR) image containing fallen stems. The
ground sampling distance of 10 cm is sufficient to delineate each individual stem with
high precision (b).}
\label{fig:refLabeling}
\end{figure} 

\subsection{Reference data}

Two separate regions of the National Park were used in this study
(Fig.~\ref{fig:natParkAreas}). We manually labeled individual stems forming
large groupings of fallen trees visible in the high-resolution aerial imagery
(Fig.~\ref{fig:refLabeling}). We only considered stems with a minimal length of
2 m. In Region A, a total of 213 single stem polygons were labeled. These
polygons formed the basis for training the semantic segmentation component
(U-net, see Sec.~\ref{sec:trainUnet}). Additionally, we used Region B,
disjoint from Region A, to derive a total of 730 fallen tree polygons
distributed across 3 test areas (Fig.~\ref{fig:natParkAreas}). We took care to
mark all visible fallen stems in each respective area to enable a fair
evaluation. The areas B1, B2, and B3 are ordered by an increasing, subjective
degree of segmentation difficulty. The first test area (B1) comprises 157 fallen
stems and a number of standing dead trees in a state of advanced decay
(Fig.~\ref{fig:testAreaB1}), distributed over an area of 140 x 70 $m^2$. The
fallen trees are often  Area B2 is slightly larger with dimensions of 140 x 70
$m^2$, but contains significantly more stems with a count of 218. It contains
some visually more challenging scenarios of many stems intersecting at various
angles. Finally, area B3 (140 x 107 $m^2$) is the most challenging among the
test plot, with 355 fallen stems and difficult scenarios of many stems
forming complex interactions. A particularly dense region within area B3 is
depicted in Fig.~\ref{fig:denseB3}.

\begin{figure}[h!]
\centering
\begin{subfigure}[b]{1.0\columnwidth}
\centering
\includegraphics[width=1.0\columnwidth]{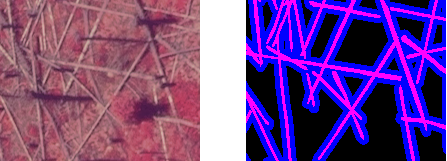}
	\caption{}
	\label{fig:labelPolyArea}
\end{subfigure}
\begin{subfigure}[b]{1.0\columnwidth}
\centering
\includegraphics[width=1.0\columnwidth]{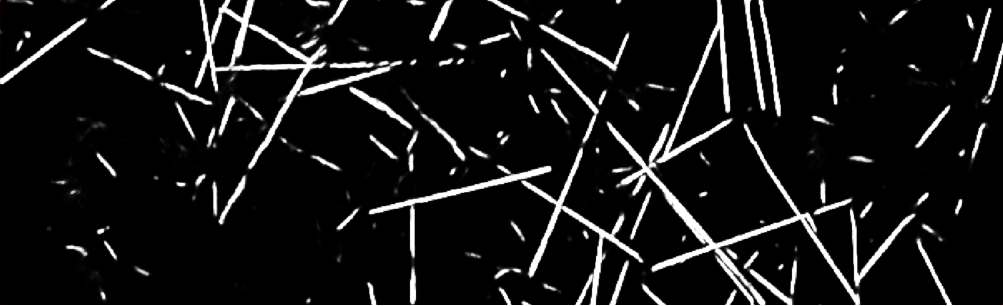}
	\caption{}
	\label{fig:labelPolyProb}
\end{subfigure}
\caption{(a) an image patch for U-net training. left: original
CIR image, right: pixel mask showing target class (magenta) and non-class
(blue) pixels. The black regions within the image do not contribute to the
training loss function, which enables the learning to focus more on the
boundary between the stem and its surroundings. (b) Per-pixel probability of
belonging to a fallen stem, obtained from semantic segmentation with the
trained U-net.}
\end{figure}

\subsection{Evaluation criteria}

We utilize two classic measures, correctness (also known as
precision/specificity) and completeness (recall/sensitivity) to quantify the
detection and segmentation results. We instantiate these measures in two
complementary settings: (i) polygon level and (ii) centerline level. In both
scenarios, correctness is conceptually defined as the ratio of detected objects
which may be linked to reference stems, whereas completeness refers to the
converse: the ratio of reference stems which have a detected counterpart. The
exact matching criteria for the polygon versus the line case are listed below.

\subsubsection{Polygon level}\label{sec:polyEval}

This version of the evaluation criteria is used for comparing the polygons
delineated by our method versus the manually created reference polygons. To
consider a reference polygon $r$ as matched, we require that there exist a
detected polygon $d$ such that $|r \cap d| / |r| > 0.5$, i.e.~more than half the
area of $r$ must be covered by the detection. Similarly, the matching criterion
for a detected polygon $d'$ is the existence of a reference polygon $r'$ such
that $|r' \cap d'| / |d'| > 0.5$. Using basic algebra of sets, it's possible to
show that these conditions taken together guarantee an intersection-over-union
(IoU) value above $1/3$. While an IoU value of 0.5 is often regarded as the
cutoff point for 'good' detection, there are several reasons why we chose a
smaller threshold: (i) the polygons represent the objects themselves, not
bounding boxes like in a typical detection scenario, (ii) some fallen stems
marked as whole within reference data may be fragmented into multiple parts by
shadows within the image, thereby making detection with a single contiguous
polygon unlikely, and (iii) the detected polygons are typically only several
pixels wide, so an error in width of merely 1 pixel could result in a
significant loss of overlap area (e.g.~15\% or more). We report the mean IoU
values on matched reference stems for relevant experiments.

\begin{figure}[h]
\centering
\includegraphics[width=1.0\columnwidth]{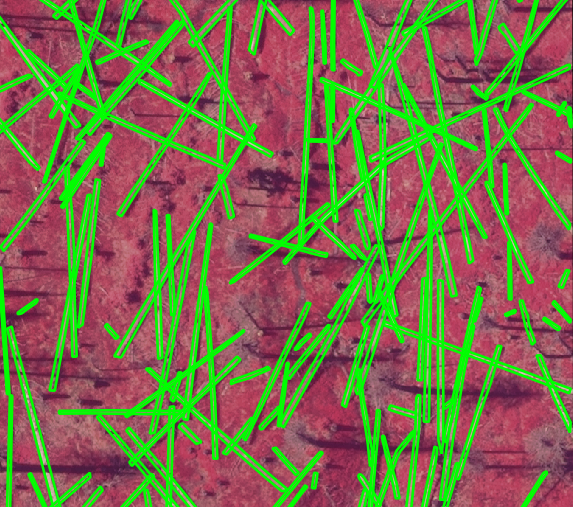}
\caption{Dense region within plot B3, where many stems are concentrated on a
relatively small area.}
\label{fig:denseB3}
\end{figure}

\subsubsection{Line level}

\begin{figure}[h]
\centering
\includegraphics[width=1.0\columnwidth]{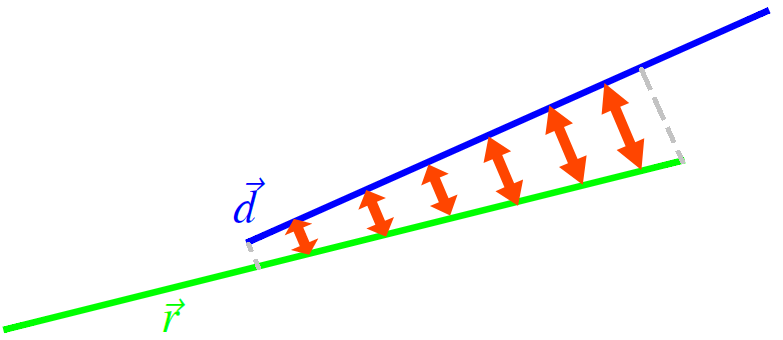}
\caption{Computing the average projection distance and cover between two line
segments $\vec{d},\vec{r}$. The average distance is taken over a discrete set of
projected points (orange distance markers). Dashed gray lines indicate the
region of $d$ covered by the projection of $r$ onto $d$.}
\label{fig:lineMatchingCriteria}
\end{figure}

Since the baseline method for comparison (i.e.~sample consensus line detection)
does not produce polygons, we introduce the line-based evaluation for the sake
of fairness. Similar to the polygon case, we perform pairwise 
comparisons between line segments extracted from the reference and detected
polygons. The segments for the reference polygons are derived from the
centerlines of their oriented bounding boxes, whereas for the detected
rectangles they are simply the centerlines parallel to the longer rectangle edge, clipped to lie within the shape. To
determine a match between two segments, we adapt the 3D matching criterion from
our prior work concerning detection of fallen stems in point clouds~\citep{Polewski2015b} to
the 2-dimensional case. Let $\vec{r},\vec{d}$ denote, respectively, the
reference and detected line segments which are candidates for matching. We
consider $\vec{r}$ matched with $\vec{d}$ if and only if the following 3
criteria are met (see Fig.~\ref{fig:lineMatchingCriteria}):
\begin{itemize}
  \item the angular deviation between $\vec{r},\vec{d}$ is
  below $5^\circ$
  \item the mean projected distance between $\vec{r},\vec{d}$ is below 35 cm,
  or half-width of the average stems we expect to encounter
  \item the projection of $\vec{r}$ onto $\vec{d}$ must have a minimum length of
  60\%$\cdot |\vec{d}|$
\end{itemize}

\begin{table*}[h!]
\centering
\ra{1.3}
\begin{tabular}{@{}lcc|cc|cc|cc||cc|cc|cc|cc@{}}\toprule
& \multicolumn{16}{c}{Collinearity coefficient}\\
& \multicolumn{8}{c}{Polygon level} & \multicolumn{8}{c}{Line level}\\
\cmidrule(l){2-9} \cmidrule(l){10-17} & \multicolumn{2}{c}{Off} &
\multicolumn{2}{c}{Low} & \multicolumn{2}{c}{Mod.} & \multicolumn{2}{c}{High} &
\multicolumn{2}{c}{Off} & \multicolumn{2}{c}{Low} &
\multicolumn{2}{c}{Mod.} & \multicolumn{2}{c}{High}\\
Precision/recall & P & R & P & R & P & R & P & R & P & R
& P & R & P & R & P & R\\\midrule
Shape coefficient & & & & & & & & & & & & & & & &\\
\textbf{Plot B1} & & & & & & & & & & & & & & & &\\
\multicolumn{1}{r}{Off}&  .90 &.81 &.90 &.80 & \textbf{.91}& \textbf{.82}& .90&
.80&.88 &.76 &.89 &.74 &\textbf{.89} &\textbf{.76} &.88 &.75\\
\multicolumn{1}{r}{Low}& .90& .81& .90& .82& -&- &- &- &.88 &.75 &.88 &.77 &- &-
&- &-\\
\multicolumn{1}{r}{Mod.}&.90 &.82 &- &- &\textbf{.91} &\textbf{.82} &- &- &.89
&.75 &- &- &.88 &.76 &- &-\\
\multicolumn{1}{r}{High}&.90 &.82 & -&- &- &- & .89 & .80 & .87 &.75 &- &- &- &-
&.86 &.73\\\midrule
\textbf{Plot B2} & & & & & & & & & & & & & & & &\\
\multicolumn{1}{r}{Off}& .91&.73 &.93 &.73 &.92 &.74 &.92 &.72 &.87 &.73 &.88
&.73&\textbf{.88} &\textbf{.75} &.87 &.73\\
\multicolumn{1}{r}{Low}& .92&.75 & .92&.74 &- &- &- &- &.87 &.75 &.87 &.73 &- &-
&- &-\\
\multicolumn{1}{r}{Mod.}&\textbf{.93} &\textbf{.75} &- &- &.92 &.75 &- &- &.87
&.74 &- &- &.87 &.74 &- &-\\
\multicolumn{1}{r}{High}&.91 &.77 & -&- &- &- &.91 &.76 &.87 &.74 & & & & &.87
&.73\\\midrule \textbf{Plot B3} & & & & & & & & & & & & & & & &\\
\multicolumn{1}{r}{Off}& .93 & .76 & .93 & .77 & .93 & .78 & .93 & .78 &.89 &.77
& \textbf{.89} &\textbf{.78} &.88 &.79 &\textbf{.89} &\textbf{.78}\\
\multicolumn{1}{r}{Low}& .93 & .77 & .92 & .78 & - & - & - & - & .88 & .76 &
\textbf{.89}&\textbf{.78} & - &- &- &-\\
\multicolumn{1}{r}{Mod.}& .93 & .78 & - & - & .91 & .78 & - & - & .88&.76 &- &-
& .88& .78&- &-\\
\multicolumn{1}{r}{High}& \textbf{.93} & \textbf{.79} & - & - & - & - &
\textbf{.93} & \textbf{.79} & .87 &.75 &- &- &- &- &.87 &.77\\
\bottomrule
\end{tabular}
\caption{Sensitivity analysis results for the influence of the shape and
collinearity energy potentials onto the aggregate energy. Precision and recall
of the detection are given for both the polygon- and line-level evaluation. Four
levels of influence are investigated for each potential, where
$\textit{'off','low','moderate','high'}$ correspond respectively to term
coefficient values of $0, 0.1, 0.3, 0.5$ within the aggregate energy. The energy term coefficient
configurations yielding the highest precision (correctness) are emphasized with bold font (recall value breaks ties).}
\label{tab:sensAnalResults_poly}
\end{table*}

\subsection{Experimental setup and results}

We performed a number of computational experiments to determine both the
absolute performance of the entire processing pipeline, its relative performance
versus a sample consensus baseline, as well as the influence of its components
on the detection quality. To facilitate computations and enable concurrent
processing, each high-probability polygon obtained from the U-net semantic
segmentation (Sec.~\ref{sec:intPrior}) is considered independently. In all
experiments, the data fit coefficient $\gamma_d$ was kept
constant at $\log \epsilon, \epsilon=1e-6$. Moreover, the overlap potential
$E_o$ is measured in the same units (i.e.~polygon area) as the data fit term, therefore we also set
$\gamma_o=\gamma_d$ to maintain the same semantics of an area unit in both
potentials.
The setting of $\epsilon$ assumes that the target class probability of pixels
outside and inside the selected image regions is respectively $1e-6, 1 - 1e-6$.
In fact, all three of the quantities $E_d,E_s,E_o$ may be interpreted as
(log-)probabilities, and we make use of this fact to define a simple potential
normalization scheme. This is to promote interpretability of the energy
coefficients $\gamma$, such that potentials having similar coefficient values
will also exert a similar influence on the energy function. In particular, we
divide each potential by the cardinality of the set it was integrated over, so
that $E_d$ is divided by the area of the currently processed high-probability
polygon $s \in S$, $E_s$ is normalized by the number of evolving shapes $M'$,
whereas the normalization constant for $E_c$ is the number of unordered pairs
$\binom{M'}{2}$.

The simulated annealing was carried out with 16 random restarts, picking the
result with the best objective function value. The number of inner iterations
per temperature level was 15000, and the cooling factor was set to 0.9. The
minimum and maximum accepted stem length was 2 m and 30 m, respectively.
Detected polygons with lengths outside this interval were discarded.

To gain a deeper insight into the experimental results, we partitioned the set
of reference trees per plot into two categories, based on their overlap with
other reference polygons. Standalone objects not intersecting any other stem
were considered 'simple', whereas stems belonging to groupings of mutually
overlapping polygons were categorized as 'complex'. The percentages of reference
trees classified as 'complex' in the test areas B1, B2, and B3 were
respectively 53\%, 59\%, and 69\%.

\subsubsection{Training the U-net}\label{sec:trainUnet}

We utilized the manually marked stem polygons to train an instance of the 3
layer U-net depicted in Fig.~\ref{fig:unet}, with additional dropout and batch
normalization layers. The implementation provided by~\citet{akeret2017radio} was
adapted to our data. The input image size $D_0$ was 200 pixels, and the number
of features (convolutional filters) at top level was set to $F=32$. Since stems are elongated thin structures, usually the
proportion of pixels occupied by them is small compared to the background.
Therefore, we only used pixels lying within a small 4-pixel band around the
marked stem polygons in the role of negative class examples. This was to
enhance the class balance and also encourage the learning process to focus on
learning the boundaries between stems and their immediate surroundings instead
of random background patterns (Fig.~\ref{fig:labelPolyArea}). The resulting
class label distribution was imbalanced with 31\% of pixels representing fallen
stems. A sample result of the probabilistic output obtained from semantic segmentation
with the trained U-net is shown in Fig.~\ref{fig:labelPolyProb}. We used the
Adam algorithm~\citep{kingma2017adam} to perform stochastic gradient
optimization of a binary logistic objective until convergence. Standard
metaparameters for the Adam optimizer were assumed ($\alpha=0.001,
\beta_1=0.9,\beta_2=0.999$). The dropout rate was 50 \%, whereas the training
minibatch size was set to 15.

\subsubsection{Sensitivity analysis for energy coefficients}

In the first experiment, we varied the energy coefficients $\gamma_s, \gamma_c$
corresponding respectively to the shape and collinearity energy terms $E_s,E_c$
(Secs.~\ref{sec:shapePriorSpecific},\ref{sec:collin}). Thanks to the
normalization scheme described above, it suffices to investigate coefficient values of the
order of magnitude $1$. We introduced 4 levels of coefficient magnitude:
$(0,0.1,0.3,0.5)$, corresponding to labels of $L=$\emph{off,low,moderate,high}.
Performance metrics were collected for the following combinations of $(\gamma_s,
\gamma_c)$: $\{(\text{off},\text{off})\} \cup
\{\{(\text{off},x),(x,\text{off}),(x,x)\}: x \in L\}$. For the polygon-based
evaluation, we recorded the correctness and completeness as per
Sec.~\ref{sec:polyEval} as well as the mean matched intersection-over-union
measure. In case of line-based evaluation, the metrics saved were (i) the
correctness and (ii) a version of completeness which considers only reference
stems which were covered by detected segments (in the projection sense, see
Fig.~\ref{fig:lineMatchingCriteria}) to a degree of at least 65\%. The results
are summarized in
Table~\ref{tab:sensAnalResults_poly}. Also,
Figs.~\ref{fig:resultsPolyB1}-\ref{fig:resultsPolyB3} visualize the detection
results of the best performing parameter combinations for the 3 target
areas. On the polygon level, a correctness above 0.9 was reached for all plots,
with completeness values between 0.77 and 0.82. The
highest attained intersection-over-union was 0.59, 0.55, and 0.58 respectively
for plots B1, B2, B3. The plot exhibiting the highest completeness was also the
one with the highest percentage of 'simple' (single component) reference trees.
Adjusting the shape and collinearity term coefficients yielded an improvement in
precision/recall of 1/1, 2/4, and 0/3 percentage points (pp) respectively for
plots B1, B2, B3. Moreover, all results with the highest attained correctness
were associated with a 'moderate' or higher shape term coefficient. In contrast,
varying coefficients did not influence the line level evaluation much, with
precision/recall gains of 1/0, 1/2, and 0/1 pp. Overall, relative to the
polygon level, the line evaluation resulted in slightly lower values for
precision at 88-89 and completeness of 75-78.

\begin{figure}[h!]
\centering
\begin{subfigure}[b]{1.0\columnwidth}
\centering
\includegraphics[width=1.0\columnwidth]{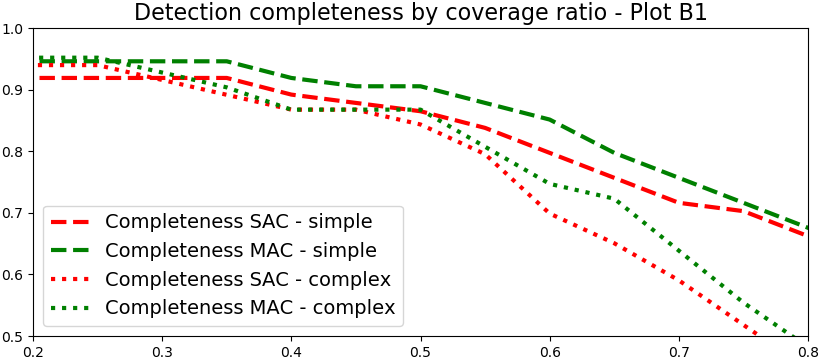}
	\caption{}
\end{subfigure}
\begin{subfigure}[b]{1.0\columnwidth}
\centering
\includegraphics[width=1.0\columnwidth]{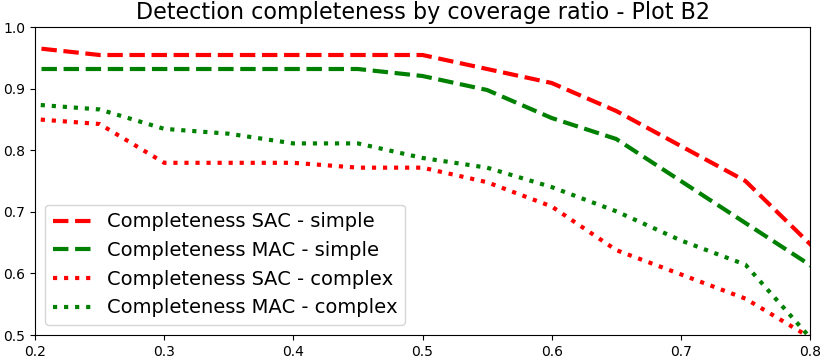}
	\caption{}
\end{subfigure}
\begin{subfigure}[b]{1.0\columnwidth}
\centering
\includegraphics[width=1.0\columnwidth]{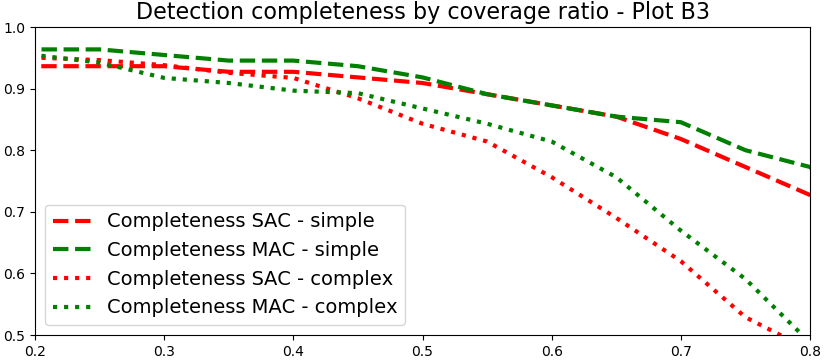}
	\caption{}
\end{subfigure}
\caption{Detection completeness results for the 3 test plots - comparison
between the sample consensus baseline (SAC) and our multiple active contour
(MAC) method.
The horizontal axis indicates the ratio of the reference tree's length which is covered by the
projection of its matched detected line. A point $(p,q)$ on the plot is
interpreted as $q$ of all reference trees having a valid match which covers at
least $p$ of their length.}
\label{fig:resultsCompletenessBaseline}
\end{figure}

\subsubsection{Comparison to baseline (sample consensus)}

The purpose of this experiment was to compare the line-based detection quality
to the sample consensus baseline. To this end, we executed the sample-consensus
based line segment detection within the high-probability components
from U-net semantic segmentation (Sec.~\ref{sec:sampleConsInit}), and considered
the SAC line segments as the final detection result. To account for
randomness and to equalize the chances versus the compared-to method, the SAC
computations were repeated a number of times equal to the size of checked
coefficient combination set in the first experiment, and the best result was
noted. We then picked the best-performing coefficient combination
per plot as per Table~\ref{tab:sensAnalResults_poly} and performed a more
in-depth comparison of our method and the SAC result using more metrics. Notably, we analyzed
completeness at different thresholds of stem coverage as well as correctness of
detecting 'simple' stems (which occupy their individual input polygon, without
intersecting other trees) versus 'complex' stems (which are part of a complex
aggregate of multiple overlapping objects). The curves showing detection completeness as a function of
reference stem projected coverage ratio are depicted in
Fig.~\ref{fig:resultsCompletenessBaseline}, whereas the remaining metrics are
given by Table~\ref{tab:resultsSac}. In terms of overall precision
(correctness), our method attains a lead of 4 pp consistently across the test
plots. However, considering the complexity of the reference stems, this
difference is extended to 5-7 pp for complex stems and reduced to 0-3 pp for
simple (single component) stems. The detection completeness (recall) follows a
similar trend, with our method outperforming the baseline by up to 5 pp for
simple and up to 7 pp for complex stems. Note that the advantage of our method
becomes more clear at coverage levels beyond 60\%, whereas for low coverage
levels, both methods perform similarly.

\subsubsection{Comparison to semantic segmentation baseline - logistic
regression}

\begin{figure}[h!]
\centering
\begin{subfigure}[b]{1.0\columnwidth}
\centering
\includegraphics[width=1.0\columnwidth]{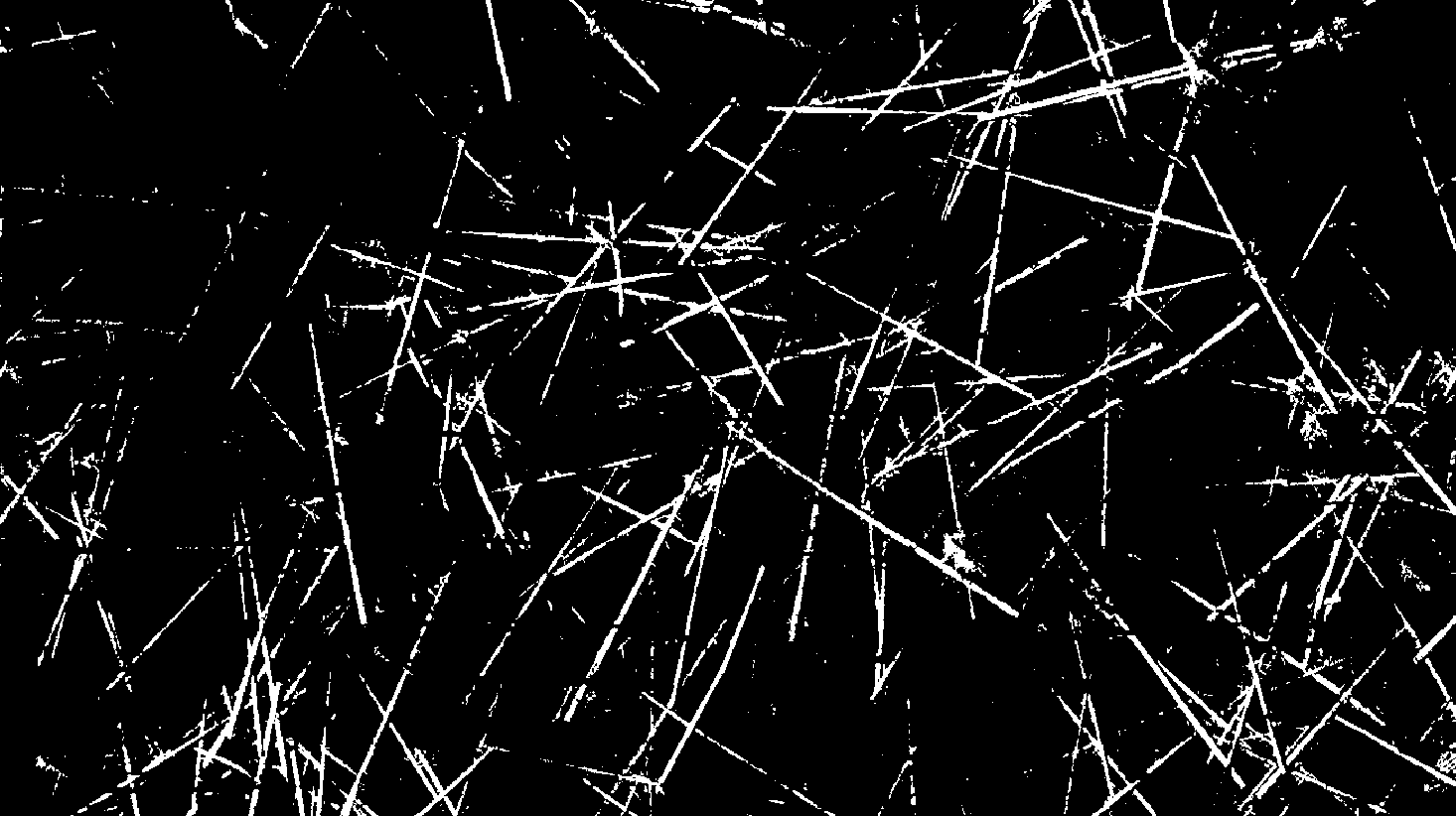}
	\caption{}
\end{subfigure}
\begin{subfigure}[b]{1.0\columnwidth}
\centering
\includegraphics[width=1.0\columnwidth]{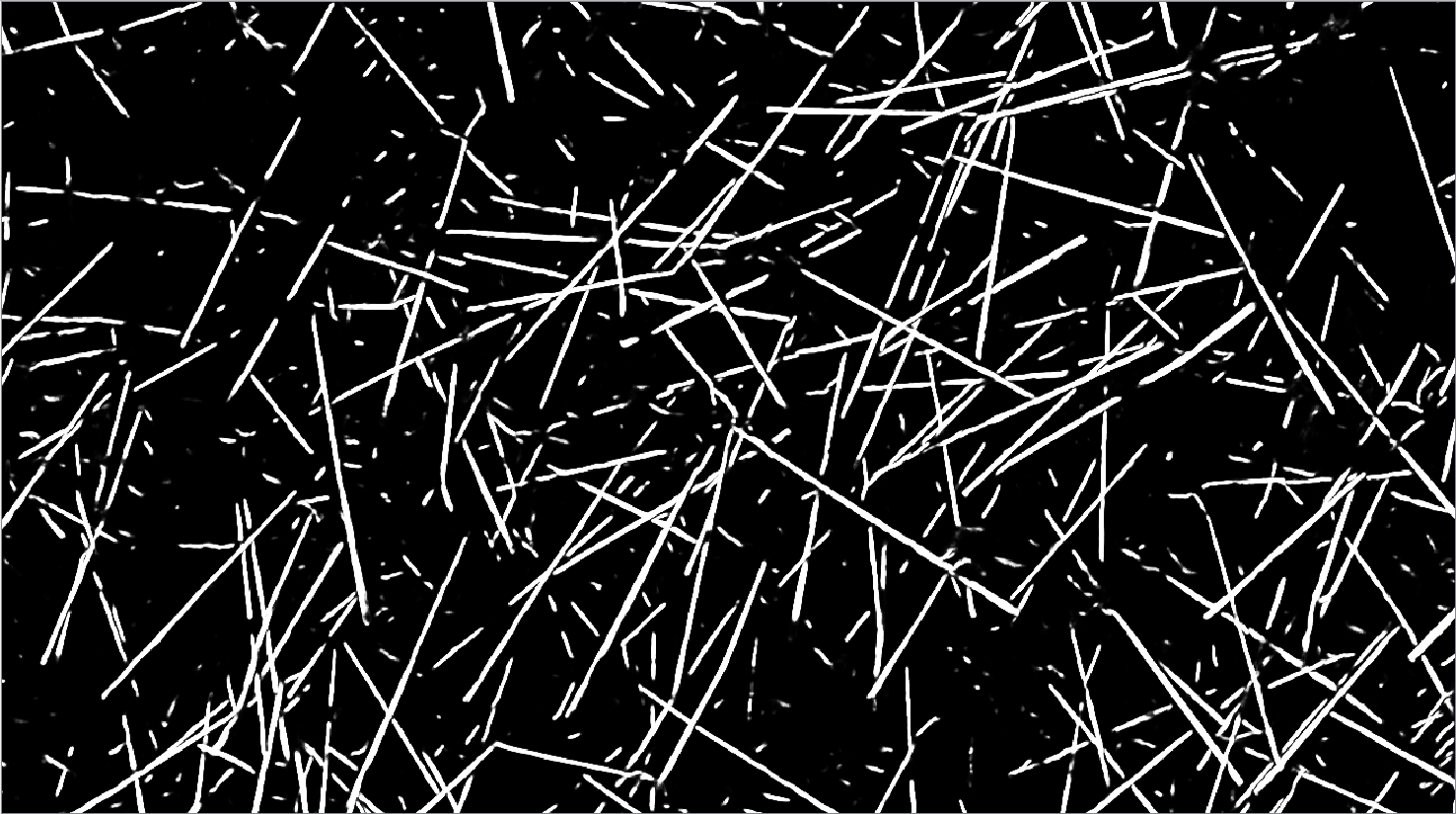}
	\caption{}
\end{subfigure}
\caption{Comparison of posterior probabilities from semantic segmentation by
(a) logistic regression based on simple channel intensities and (b) U-net. The
LR baseline tends to thin out the stems, often reducing them to sparse sets of
pixels.}\label{fig:cmpLRprobs}
\end{figure}

This experiment involved replacing the high-quality semantic segmentation
probability map from the U-Net with a basic logistic regression model trained
only on the channel intensities. The same data was used for training both
models. No higher-level textural features were used in order to determine the
benefit of using a state-of-the-art neural network for semantic segmentation. The pixel-level classification accuracy and F1 score on a
hold-out validation set for the U-Net were respectively 0.94 and 0.91. The
cross-validated overall accuracy and F1 score for the logistic regression
baseline attained values of 0.80 and 0.62. We then applied the logistic
regression model to the images of the test area, obtaining maps of posterior
class probabilities. Polygons of high
probability regions were subsequently extracted and our multi-contour
segmentation was executed. Although the per-pixel classification metrics for the
logistic regression were satisfactory, the object-level (both line and polygon)
performance appeared to break down. The precision degraded to levels of
0.76-0.83, and the recall experienced an even more extreme drop to levels of
0.15-0.27. In Fig.~\ref{fig:cmpLRprobs}, the semantic segmentation of the same
area by the LR baseline and by the U-Net is shown. It can be seen that within
the LR probability image, many stems are missing or greatly 'thinned out', i.e.
represented by only a sparse set of pixels.

\subsection{Execution time}

The training process of the U-net on an nVidia GeForce GTX 1080 Ti graphics card
(with CUDA support) took ca.~7.5 hours, after which time convergence of the
learning process was achieved. The prediction time of the U-net on new data was
measured in seconds and negligible compared to the optimization time of the
multi contour objective. This optimization was carried out on a desktop computer
equipped with 128 GB of RAM and an Intel XEON E5-1680 v4 CPU running at a
frequency of 3.4 GHz, consisting of 8 cores. We used our own implementation of
the simulated annealing metaheuristic algorithm written in the C++ programming
language. The mean execution times of the inference/optimization on the
respective test areas (averaged over different choices of the objective function
parameters $\gamma_s, \gamma_c$) are given in Table~\ref{tab:execTimes}.

\begin{table}[h!]
\centering
\ra{1.3}
\begin{tabular}{@{}lccc@{}}\toprule
& Plot B1 & Plot B2 & Plot B3\\\midrule
Mean exec.~time [h] & 3.95 & 6.68 & 13.46\\
Standard dev.~[h] & 0.17 & 0.42 & 0.46\\
Time per stem [s] & 61 & 83 & 91\\
\bottomrule
\end{tabular}
\caption{Execution times of simulated annealing based optimization of proposed
multi active contour method on the 3 test areas. The shown values correspond to
the mean execution time, standard deviation, and mean time for processing one
stem per test area.}
\label{tab:execTimes}
\end{table}

\begin{table*}[h!]
\centering
\ra{1.3}
\begin{tabular}{@{}lcccccc@{}}\toprule
& Pr.(total) & Pr.(simple) & Pr.(complex) & Rec.(total)\\\midrule
\textbf{Plot B1} & & & &\\
\multicolumn{1}{r}{SAC}&  .85 &.82 &.90 &.70\\
\multicolumn{1}{r}{MAC}&  .89 &.85 &.97 &.76\\\midrule
\textbf{Plot B2} & & & &\\
\multicolumn{1}{r}{SAC}&  .84 &.89 &.79 &.73\\
\multicolumn{1}{r}{MAC}&  .88 &.91 &.84 &.75\\\midrule
\textbf{Plot B3} & & & &\\
\multicolumn{1}{r}{SAC}&  .84 &.88 &.82 &.74\\
\multicolumn{1}{r}{MAC}&  .88 &.88 &.87 &.79\\
\bottomrule
\end{tabular}
\caption{Results of line-based evaluation - comparison between baseline sample
consensus (SAC) and our multiple active contour (MAC) method. Shown are the
precision (Pr.) on the whole data, for 'simple' and for 'complex' reference
stems, as well as the total recall (Rec.) at 0.65 coverage of reference stems. }
\label{tab:resultsSac}
\end{table*}

\begin{figure*}[ht]
  \centering
  \begin{subfigure}[b]{0.9\textwidth}
  \centering
	\includegraphics[width=1.0\textwidth]{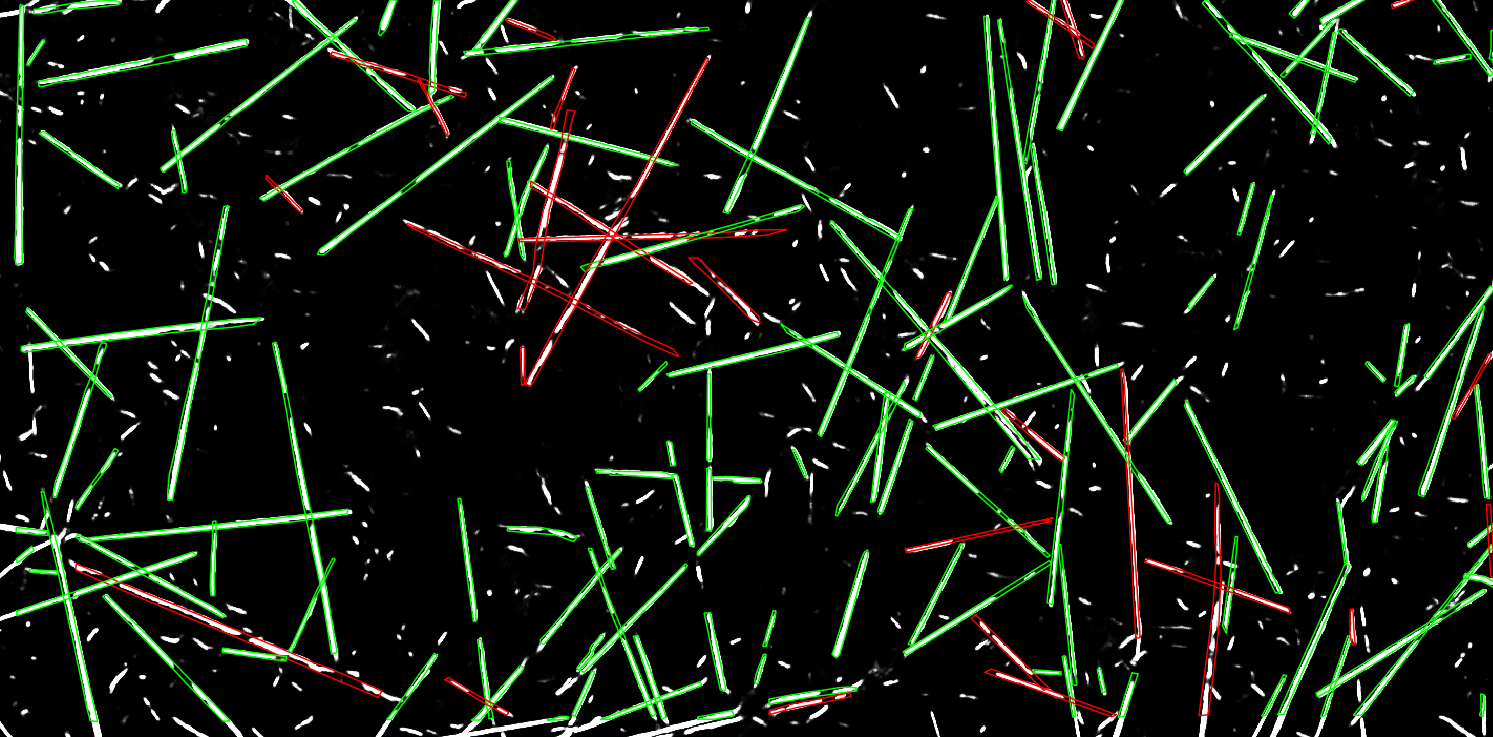}
	\caption{}
	\label{fig:polyRes_B1_ref}
\end{subfigure}
\begin{subfigure}[b]{0.9\textwidth}
\centering
\includegraphics[width=1.0\textwidth]{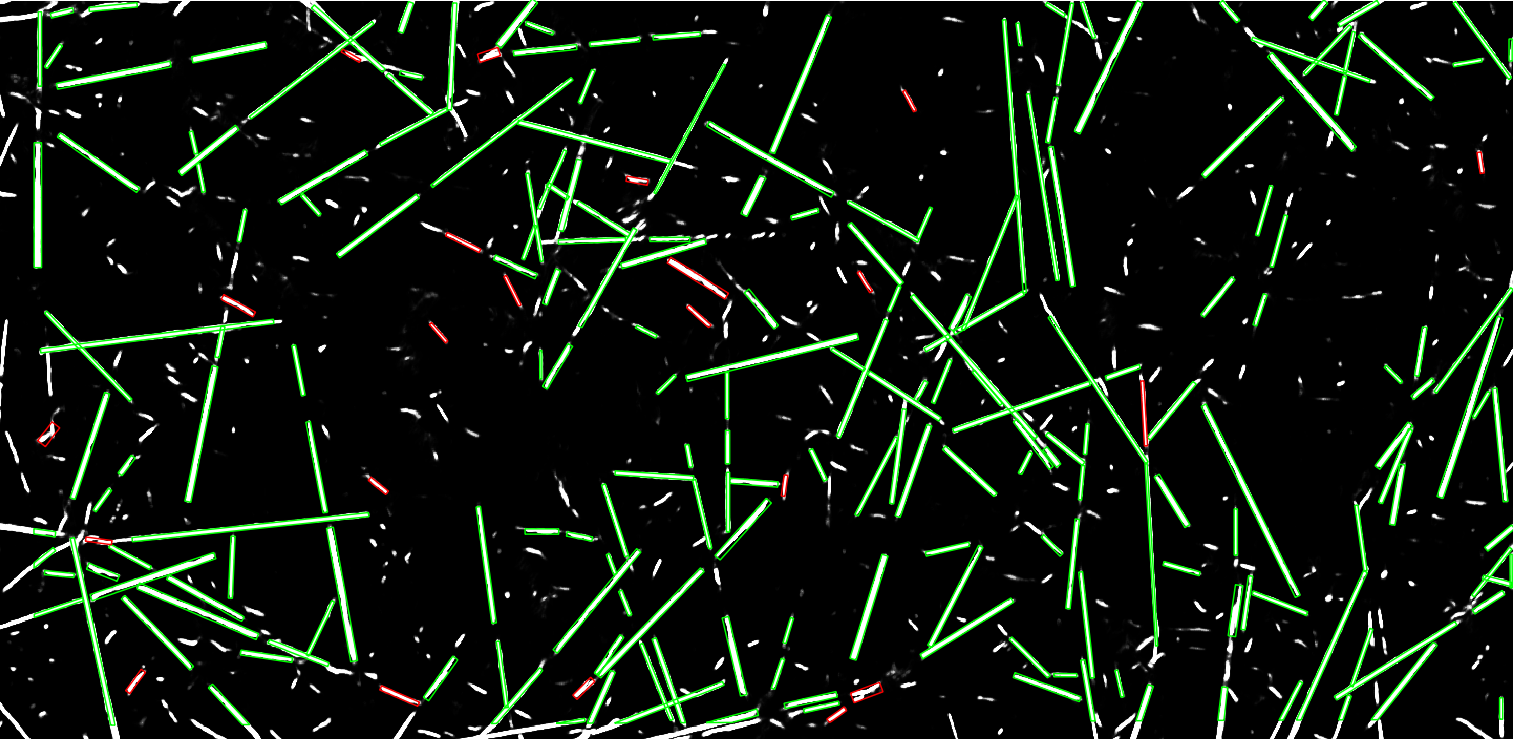}
	\caption{}
	\label{fig:polyRes_B1_det}
\end{subfigure}
\begin{subfigure}[b]{0.9\textwidth}
\centering
\includegraphics[width=1.0\textwidth]{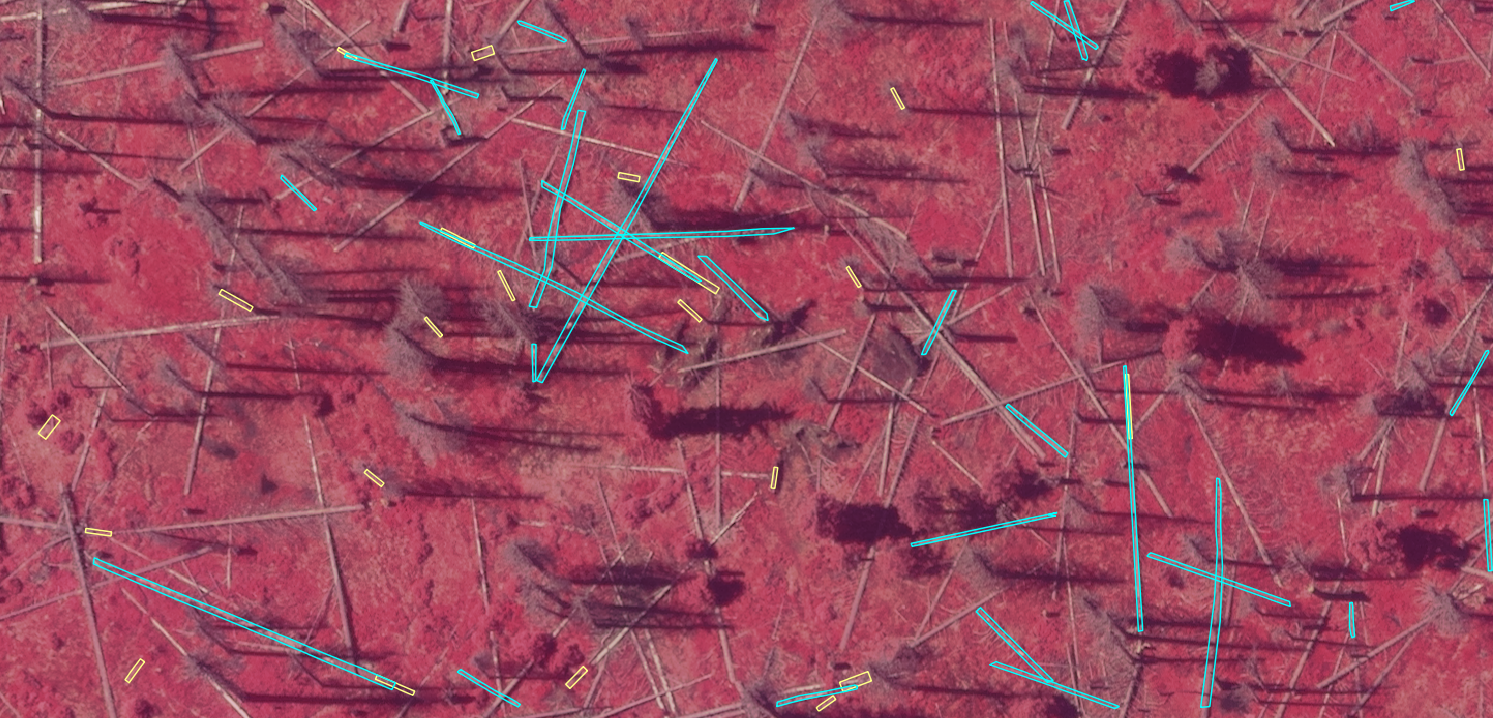}
	\caption{}
	\label{fig:polyRes_B1_cir}
\end{subfigure}
\caption{Results of fallen stem segmentation for plot B1 (polygon level). (a),
(b) depict respectively the reference and detected polygons, with semantic
segmentation posterior probability as background. Red/green colors indicate a
polygon mismatched/matched with a counterpart (above 50\% area overlap). (c)
original false color CIR image with indicated mismatched reference (cyan) and
detected (yellow) polygons.}
\label{fig:resultsPolyB1}
\end{figure*}

\begin{figure*}[ht]
  \centering
  \begin{subfigure}[b]{0.7\textwidth}
  \centering
	\includegraphics[width=1.0\textwidth]{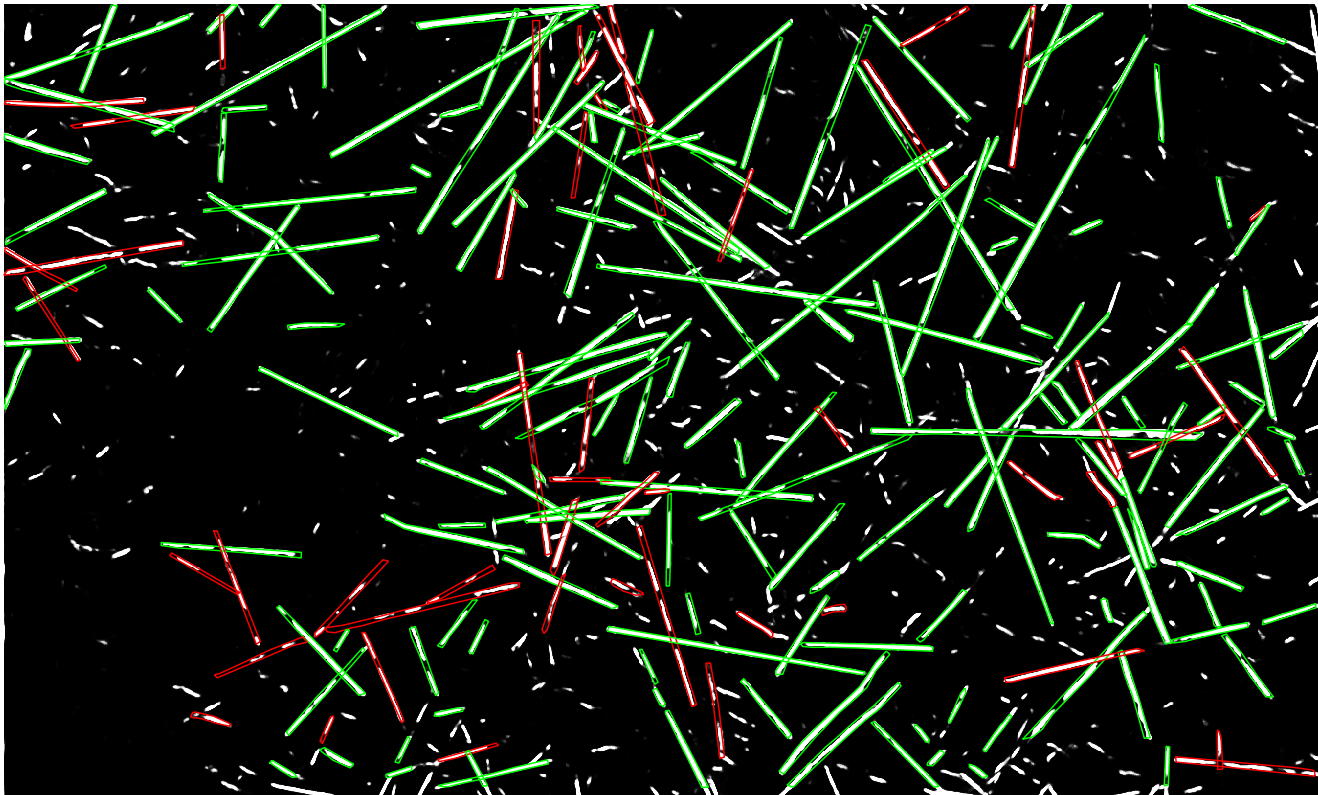}
	\caption{}
	\label{fig:polyRes_B2_ref}
\end{subfigure}
\begin{subfigure}[b]{0.7\textwidth}
\centering
\includegraphics[width=1.0\textwidth]{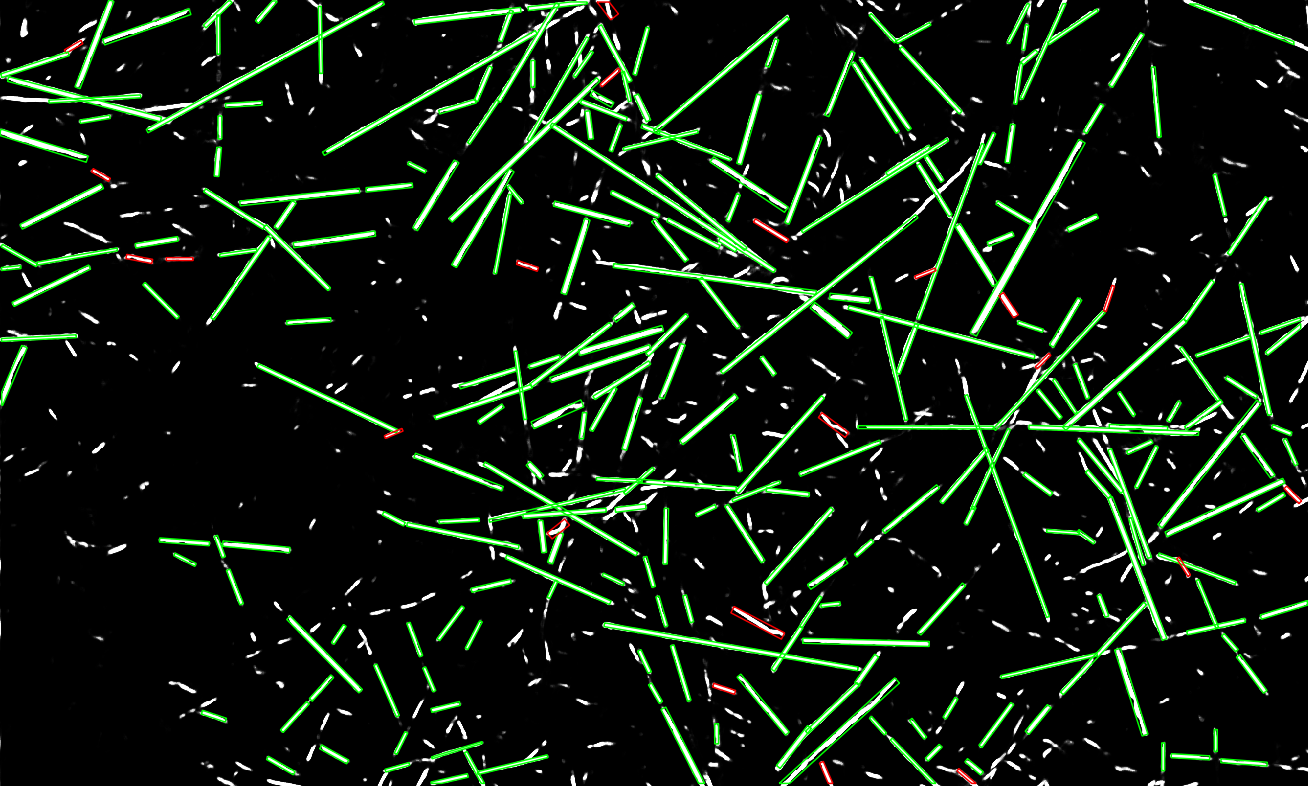}
	\caption{}
	\label{fig:polyRes_B2_det}
\end{subfigure}
\begin{subfigure}[b]{0.7\textwidth}
\centering
\includegraphics[width=1.0\textwidth]{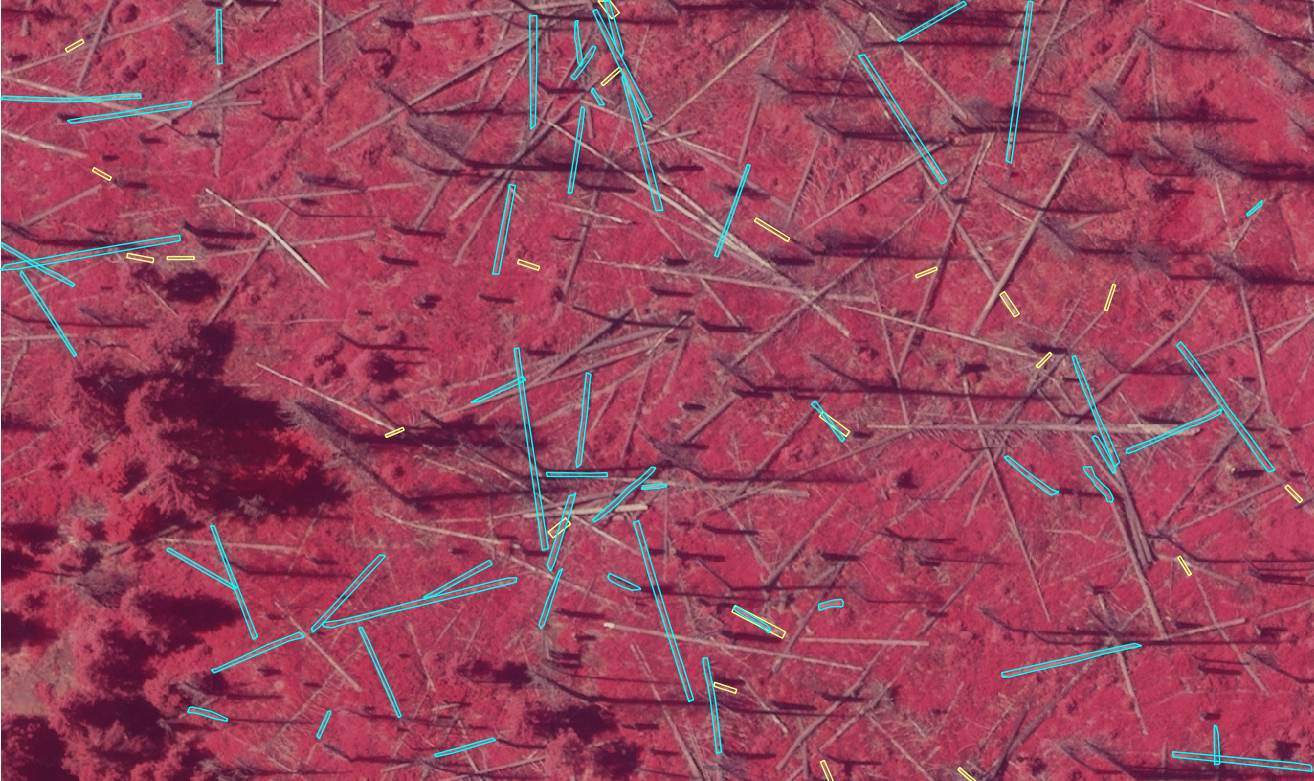}
	\caption{}
	\label{fig:polyRes_B2_cir}
\end{subfigure}
\caption{Results of fallen stem segmentation for plot B2 (polygon level). (a),
(b) depict respectively the reference and detected polygons, with semantic
segmentation posterior probability as background. Red/green colors indicate a
polygon mismatched/matched with a counterpart (above 50\% area overlap). (c)
original false color CIR image with indicated mismatched reference (cyan) and
detected (yellow) polygons.}
\label{fig:resultsPolyB2}
\end{figure*}

\begin{figure*}[ht]
  \centering
  \begin{subfigure}[b]{0.49\textwidth}
  \centering
	\includegraphics[width=1.0\textwidth]{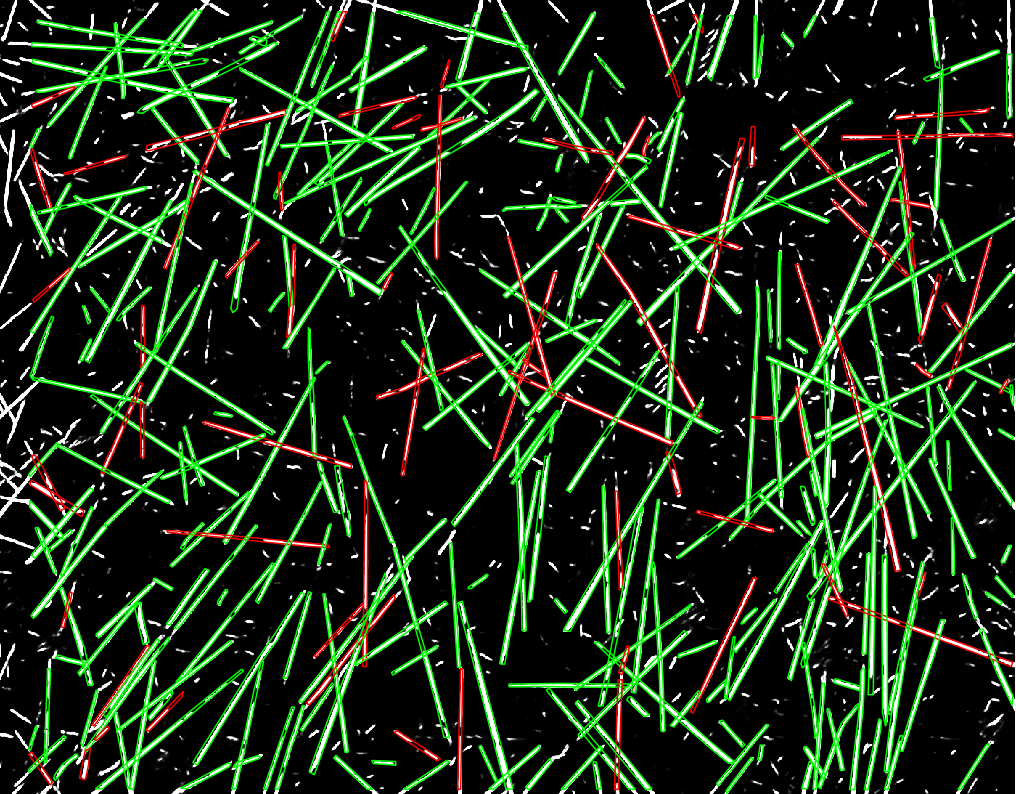}
	\caption{}
	\label{fig:polyRes_B3_ref}
\end{subfigure}
\hspace{\fill}
\begin{subfigure}[b]{0.49\textwidth}
\centering
\includegraphics[width=1.0\textwidth]{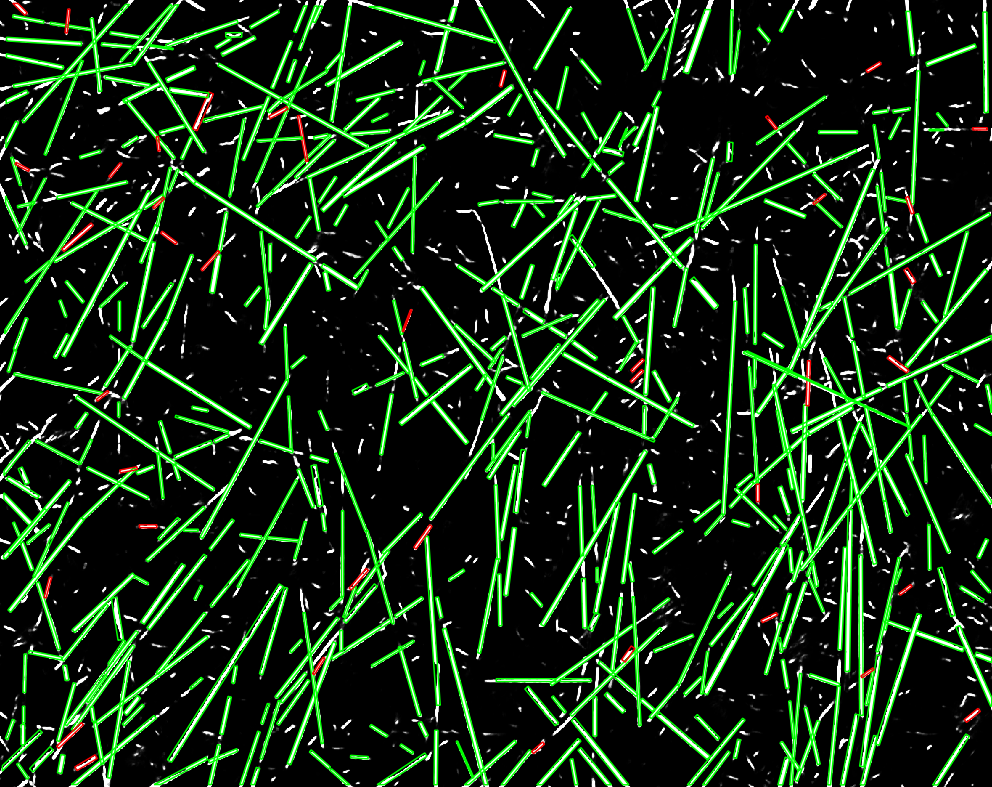}
	\caption{}
	\label{fig:polyRes_B3_det}
\end{subfigure}
\begin{subfigure}[b]{0.7\textwidth}
\centering
\includegraphics[width=1.0\textwidth]{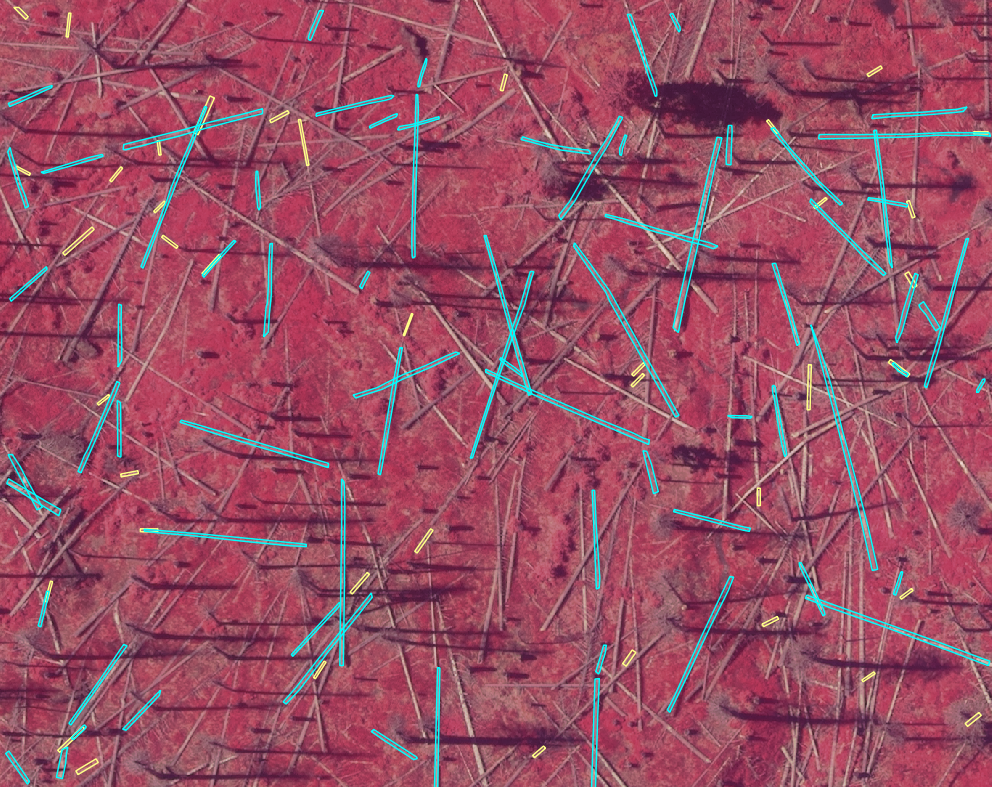}
	\caption{}
	\label{fig:polyRes_B3_cir}
\end{subfigure}
\caption{Results of fallen stem segmentation for plot B3 (polygon level). (a),
(b) depict respectively the reference and detected polygons, with semantic
segmentation posterior probability as background. Red/green colors indicate a
polygon mismatched/matched with a counterpart (above 50\% area overlap). (c)
original false color CIR image with indicated mismatched reference (cyan) and
detected (yellow) polygons.}
\label{fig:resultsPolyB3}
\end{figure*}

\section{Discussion}\label{sec:discussion}

Overall, our method was successful in providing a good quality detection result
for all 3 test plots of multi-level scenario complexity, both in terms of
agreement of the extracted and reference polygons (IoU between 0.55-0.59), and the percentage of matched reference and
detected polygons (correctness of 0.91-0.93, completeness 0.78-0.82). As
expected, the shape prior turned out to be more helpful in case of polygon
level evaluation, because the line level evaluation is less sensitive to changes
in detected polygon width and small changes in orientation. Despite the
simplicity of the utilized shape representation (rectangles parameterized by
width and length), the energy benefited from an explicit shape model with a gain
of up to 4 pp in completeness (while maintaining correctness). We hypothesize
that a more complex shape model could show even higher gains. The test plot B1,
which benefited the least from the additional energy terms, also had the lowest
percentage of complex (intersecting) reference stems, confirming the intuition
that the shape and collinearity priors are mostly useful for the complex
scenarios.

It is interesting to note that plot B1, which can be considered the 'easiest',
obtained the lowest precision score among the 3 test plots on the polygon level.
This can be attributed to a relatively high number of standing dead tree stems within this plot (see Fig.~\ref{fig:resultsPolyB1}c). These stems appear to be virtually
indistinguishable from lying stems under the semantic segmentation output of the
U-Net. This is probably a consequence of the network not being trained to
distinguish standing dead trees from fallen stems. It is not clear whether this
can be achieved solely based on monocular images without dense depth
information. Aside from standing dead trees, other sources of false negatives
may be linked to root plates as well as woody debris appearing to possess a
similar hue within the CIR images as our target objects.

A number of misdetections (unmatched reference trees) is once again associated
with the posterior probability of the semantic segmentation from the U-Net. As
visible on
Figs.~\ref{fig:resultsPolyB1}a,~\ref{fig:resultsPolyB2}a,~\ref{fig:resultsPolyB3}a,
the missing stems are often fragmented into discontinuous chunks in the
probability image, caused mostly by shadows and occlusions from
other objects like shrubs or understory growth. Such discontinuities prohibit
the energy function from enclosing the disjoint stem parts in a single detected
polygon. This is associated with our method's inherent tendency to exploit the
connectivity structure of the high-probability pixels, where each connected
component is processed independently. Due to computational tractability
considerations, for large scenes it is impractical to consider all connected
components within one, simultaneous optimization problem. However, there are
several alternative possibilities of alleviating this problem.
First, explicitly adding examples of shadowed fallen stems to the U-Net's training set would help
increase the continuity of the stems within the probability image. Second, out
energy formulation could be altered to explicitly account for these
discontinuous, collinear detections. Finally, a post-processing step could be
applied, where the detected polygons would be clustered together based on mutual
distance and collinearity, for example using graph cut methods~\citep{Shi2000}.
In our setting, the collinearity potential from Sec.~\ref{sec:collin} could be
directly used in the role of the object similarity function.

Comparison to the sample consensus baseline shows that the line-based detection
can be improved by applying our energy function to the SAC candidates, both in
terms of completeness and correctness. Moreover, our method yields higher gains
for more complex scenarios of intersecting stems. In case of simple,
single-component stems, sample consensus line fitting usually delivers good
results and is difficult to significantly improve upon. Also, it appears that
SAC is often able to provide low-coverage partial matching of the majority of
stems present within the test area, but falls short of the task of precisely
delineating their extents. It is for higher stem length coverages that our
multiple active contour method, endowed with prior knowledge about the size and spatial
conformation of fallen stems, is able to gain the most clear advantage.

Our results show the importance of using a high-quality semantic segmentation
method as a basis for the contour evolution. We believe that the performance
degradation turned out to be so extreme because of the nature of the classified
objects. Indeed the fallen stems are usually represented by objects of only a
few pixels of width, and therefore the deformations caused by the lower-quality
logistic regression semantic segmentation turned out to distort the appearance
of stems in the probability image beyond recognition. It was nevertheless
surprising that a ca.~20\% drop in pixel-level accuracy resulted in a nearly
60\% degradation in object-level recall.

The relative execution times are consistent with our a priori ordering of the
three test areas with respect to their difficulty. Indeed, the unit time
required for processing one stem in Plots B2 and B3 is respectively 33 \% and 50
\% larger compared to Plot B1 (see Table~\ref{tab:execTimes}). The processing
time is dominated by solving the multiple active contour evolution objective
(via simulated annealing), with the semantic segmentation with the U-net as well
as the sample consensus-based line segmentation contributing only a small fraction of time. In
turn, the simulated annealing algorithm's computational complexity can be traced
back to the complexity of the move-making procedure, which is directly
proportional to the number evolving model shapes as well as the number of points forming 
the connected component's polygon (see
Sec.~\ref{sec:energyComponents}). Therefore, a single connected component with
a very complex boundary (e.g.~Fig.~\ref{fig:denseB3}) can dominate the the
processing time, especially if the initial sample consensus line segmentation
results in many model shapes to evolve. The current processing times on a
single machine are satisfactory for small and medium-scale applications of areas
which are densely covered with fallen stems. However, in this study our primary focus
was to attain high accuracy of the stem delineation and not as much to optimize
the throughput of the computation. In particular, we did not conduct
investigations into the minimal required random restarts of the simulated
annealing runs, the number of iterations of each temperature, or the cooling
schedule itself. We believe that there is potential to reduce the current
execution times by at least tenfold once these meta-parameters are optimized.
This would bring the unit cost of processing a single stem into the realm of
several seconds, which would mean that an area containing 10,000 fallen stems
could be processed within one day on a single machine.

We believe that our study showed the advantage of using active contour evolution
over generic line detection methods for the purpose of segmenting elongated
structures such as fallen tree stems in high-resolution aerial imagery. To the
best of our knowledge, it is the first study which (i) was based on more than
700 objects, (ii) provided both pixel-level as well as line-level detection
metrics, and (iii) dealt with extremely complex overlapping stem scenes. The
results show that a segmentation method which is informed on the shape and
appearance of the objects it is trying to segment can improve performance
especially in the case of complex scenes. A further advantage of our proposed
method over off-the-shelf segmentation procedures is that most of the crucial
parameters can be learned from training examples. However, it should be noted
that our study had several limitations that should be addressed in future
research. First, the presence of shadows and occlusions can be detrimental to
the formation of connected components within the posterior probability image,
leading to partition of the same physical object into multiple, unrelated
segmented objects. Moreover, the utilized rectangular representation of stems
may be too simple in some cases, especially in the context of applying the
method to more complex shapes aside from fallen stems. Also, our input data
lacked 3D information, which led to confusion between fallen and standing trees
in some cases. Finally, the meta-parameters of the simulated annealing optimizer
were not tuned for efficiency of processing, which makes the current version of
our software not applicable to large area processing. Nonetheless, we believe
that the trainable nature of the key parameters makes our approach applicable to
new, previously unseen areas given enough training data, without the need for
manual parameter tweaking.

Our results are very promising as we can count the number of fallen trees and
determine the area covered by each tree very accurately from aerial imagery.
Therefore, the proposed methodology will allow many applications in forest and
conservation management. After severe disturbances, our method allows a quick
assessment of the number and distribution of fallen trees, which is necessary
to plan salvage logging activities to harvest the timber and to prevent the
spread of insects, such as the Norway spruce bark beetle \emph{Ips typographus}.
In the next step, the delineated polygons will be a basis for determining not only
the number of the fallen trees, but also the amount of wood. This would make
the information even more suitable for forest management, since from this value
the operation of logging machinery and transportation can be planned
accurately. For conservation management, our method will help to map the
distribution and amount of deadwood in the ecosystem. This will allow to
determine the best areas for conservation measures and to monitor the amount of
lying dead wood in a given area to fulfil minimum requirements for maintaining
biodiversity~\citep{Mller2010ARO}. Moreover, our method can also be used for
research projects that need accurate information about the distribution of lying dead wood, such as
long-term studies on carbon sequestration, the spatial arrangement of forest
regeneration, or animal movement.

\section{Conclusions and outlook}\label{sec:conclusions}

This work introduced a framework for segmenting multiple objects of a common
type from imagery using a collection of evolving active contours, unified under
an aggregate energy functional encompassing various aspects of the segmentation
quality. In particular, along with the usual data fit term, our energy favors
high-probability shapes as defined by an explicit shape model, and penalizes
overlap of adjacent contours. The proposed approach makes use of
state-of-the-art semantic segmentation methods (e.g.~U-net) to extract regions
of the input image which are likely to contain realizations of target class
objects. We then instantiated the framework in the context of fallen tree
detection from high-resolution aerial color infrared imagery, by providing
concrete shape parametrizations, a kernel density estimator-based shape model,
as well as additional, domain-specific energy potentials. It was shown on 3 test
plots that our approach can achieve good segmentation performance in terms of
both polygon-based (intersection-over-union) and line-based quality metrics. It
was found that using the proposed shape model improved the segmentation
completeness at polygon level by up to 4 pp. As expected, the additional energy
terms (collinearity and shape model) were mostly useful for complex aggregates
of multiple overlapping stems, while their impact on isolated stem detection was
minimal.

Our investigation showed the critical importance of using a high-quality
semantic segmentation method in case of thin, elongated objects such as fallen
stems. The posterior probability map obtained from a simple baseline using
channel intensities resulted in a breakdown of segmentation completeness, with
many stems under-represented and fragmented in the probability image. The
sufficient quality of the semantic segmentation is a precondition for the
successful application of our method.

On the line level, the proposed energy-based segmentation method was compared to
a sample consensus baseline. Although the energy functional evolves polygons
(contours), an improvement in line-based metrics was also observed, with gains
in both precision and recall up to 6 pp.

An issue to be addressed in future work is associated with objects split by
shadows or occlusions in the probability image, leading to fragmentations of
stems into disjoint parts. For large scenes with hundreds or thousands of
objects, it would be computationally intractable to jointly consider all
high-probability image regions within one optimization problem. Instead, a more
feasible strategy seems to perform merging of the detected polygons as a
post-processing step, e.g.~using a graph-cut approach. Another natural direction
for future work is the application of our framework to more complex object
classes and associated, richer shape models. Also, instantiating the framework
in 3D using outputs from voxel-based deep semantic segmentation networks could
be an interesting next step.

\section*{Acknowledgments}
The work described in this paper was substantially supported by a grant from the Research Grants Council of the Hong
Kong Special Administrative Region, China (Project No. PolyU
25211819) and was partially supported by grants 1‐ZE8E and G-YBZ9 from The Hong Kong Polytechnic University.


%\section*{References}
\label{sec:references}
\bibliographystyle{model2-names}\biboptions{authoryear} 
\bibliography{asmIsprs}


%\begin{thebibliography}{00}



%\end{thebibliography}

\end{document}